\documentclass[final]{cvpr}

\usepackage[toc,page]{appendix}
\usepackage{times}
\usepackage{epsfig}
\usepackage{graphicx}
\usepackage{amsmath}
\usepackage{amssymb}
\usepackage{bbm}
\usepackage{booktabs}
\usepackage[font=footnotesize,labelfont=footnotesize,hypcap=true]{caption}
\usepackage[hypcap=true,list=true]{subcaption}
\usepackage{ctable}
\usepackage{blindtext}
\usepackage{tabularx}
\usepackage{multirow}
\usepackage{multicol}
\usepackage{makecell}
\usepackage{arydshln}
\usepackage{soul}
\usepackage{algorithm}
\usepackage[noend]{algpseudocode}

\usepackage[pagebackref=true,breaklinks=true,colorlinks,bookmarks=false]{hyperref}
\usepackage{cleveref}
\newcolumntype{C}[1]{>{\hsize=0.93\hsize\centering\arraybackslash}X}
\newcolumntype{D}[1]{>{\hsize=#1\hsize\centering\arraybackslash}X}
\newcolumntype{L}[1]{>{\hsize=#1\hsize\arraybackslash}X}
\newcolumntype{Y}[1]{>{\centering\arraybackslash\setlength\hsize{#1\hsize}}X}
\def\cvprPaperID{10479} %
\def\confYear{CVPR 2021}

\begin{document}
\title{The Lottery Ticket Hypothesis for Object Recognition}
\author{Sharath Girish\thanks{First two authors contributed equally\newline     \hspace*{15pt}To appear at CVPR 2021}\\
{\tt\small sgirish@cs.umd.edu}
\and
Shishira R. Maiya\footnotemark[1]\\
{\tt\small shishira@umd.edu}
\and
Kamal Gupta\\
{\tt\small kampta@umd.edu}
\and
Hao Chen\\
{\tt\small chenh@umd.edu}
\and
Larry Davis\\
{\tt\small lsd@umiacs.umd.edu}
\and
Abhinav Shrivastava\\
{\tt\small abhinav@cs.umd.edu}
}
\date{University of Maryland, College Park}
\maketitle

\begin{abstract}
Recognition tasks, such as object recognition and keypoint estimation, have seen widespread adoption in recent years. Most state-of-the-art methods for these tasks use deep networks that are computationally expensive and have huge memory footprints. This makes it exceedingly difficult to deploy these systems on low power embedded devices. Hence, the importance of decreasing the storage requirements and the amount of computation in such models is paramount. The recently proposed Lottery Ticket Hypothesis (LTH) states that deep neural networks trained on large datasets contain smaller subnetworks that achieve on par performance as the dense networks. In this work, we perform the first empirical study investigating LTH for model pruning in the context of object detection, instance segmentation, and keypoint estimation. Our studies reveal that lottery tickets obtained from Imagenet pretraining do not transfer well to the downstream tasks.
We provide guidance on how to find lottery tickets with up to 80\% overall sparsity on different sub-tasks without incurring any drop in the performance. Finally, we analyse the behavior of trained tickets with respect to various task attributes such as object size, frequency, and difficulty of detection. Our code is made public at: \href{https://github.com/Sharath-girish/LTH-ObjectRecognition}{https://github.com/Sharath-girish/LTH-ObjectRecognition}.
\end{abstract}

\section{Introduction}
\begin{figure}[t]
\centering
    \begin{tabular}{c}
        \footnotesize{ResNet-18 on COCO} \\
        \includegraphics[trim={0 1.5cm 0 0}, clip, width=\linewidth]{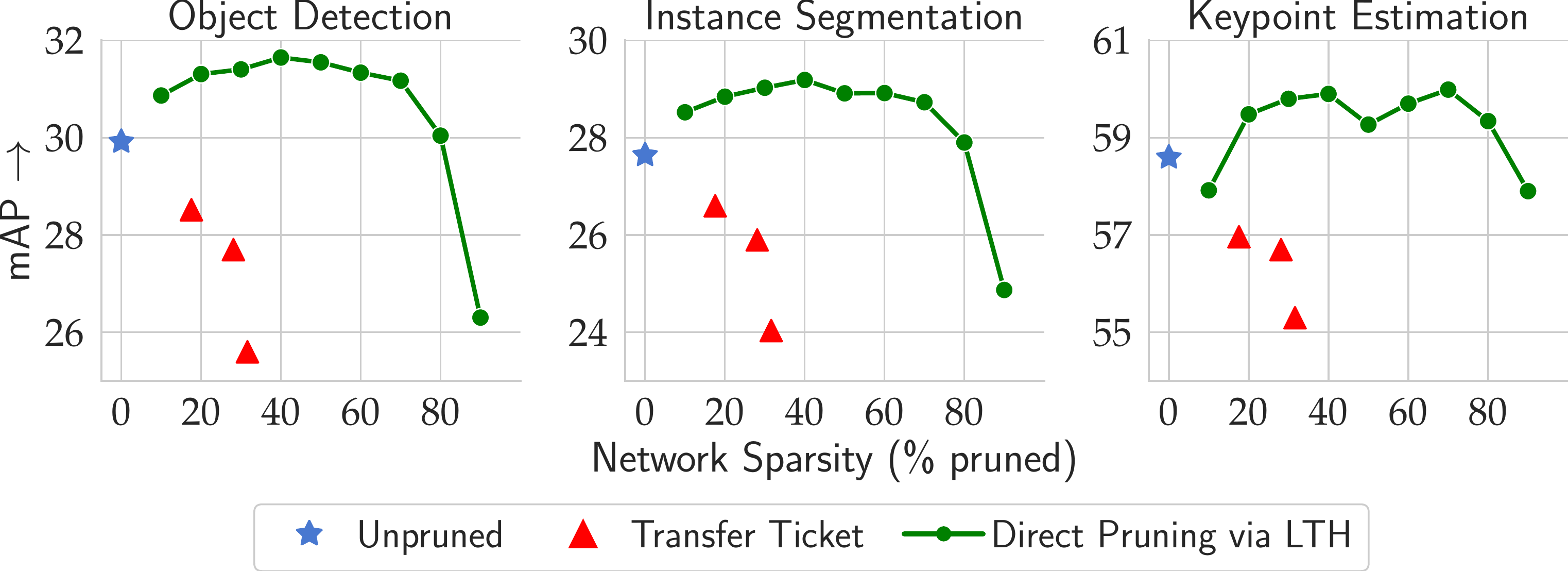}
    \end{tabular}
    \begin{tabular}{c}
        \footnotesize{ResNet-50 on COCO} \\
        \includegraphics[width=\linewidth]{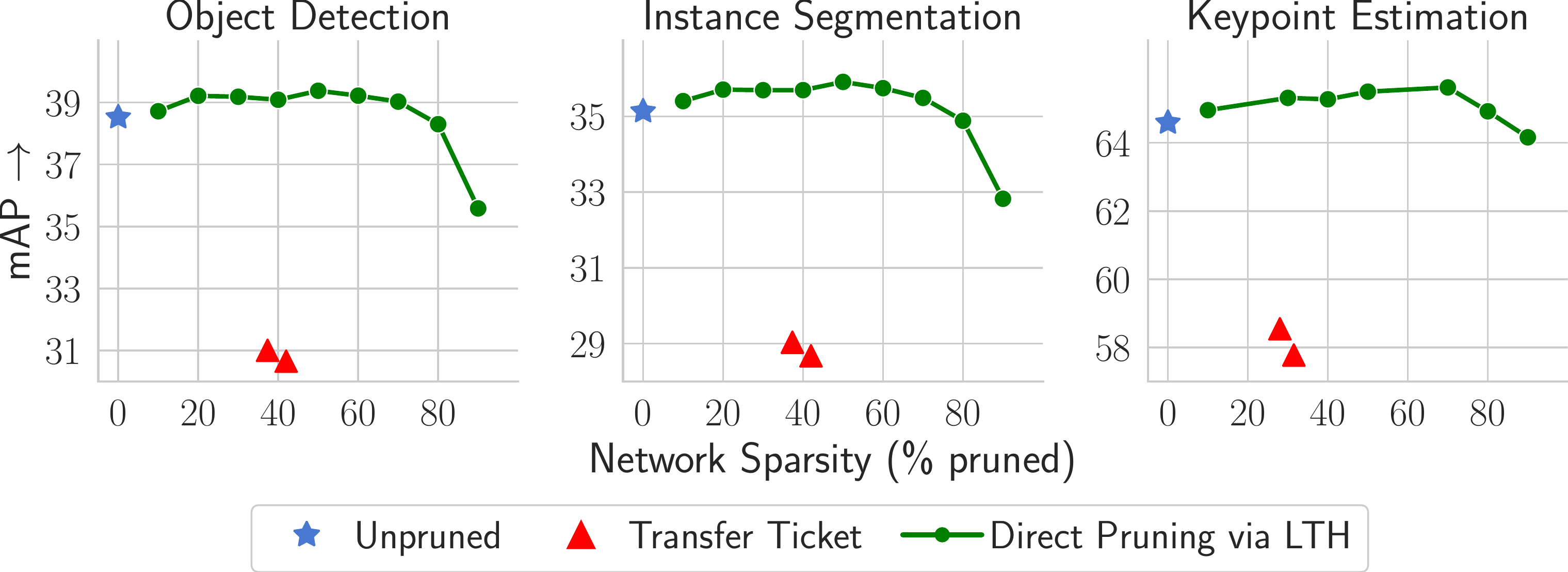}
    \end{tabular}
\vspace{-0.1in}
\caption{Performance of lottery tickets discovered using direct pruning for various object recognition tasks. Here we have used a Mask R-CNN model with ResNet-18 backbone (top) and ResNet-50 backbone (bottom) to train models for object detection, segmentation and human keypoint estimation on the COCO dataset. We show the performance of the baseline dense network, the sparse subnetwork obtained by transferring ImageNet pre-trained ``universal'' lottery tickets, as well as the subnetwork obtained by task-specific pruning. Task-specific pruning outperforms the universal tickets by a wide margin. For each of the tasks, we can obtain the same performance as the original dense networks with only 20\% of the weights.}
\vspace{-0.15in}
\label{fig:teaser}
\end{figure}

Recognition tasks, such as object detection, instance segmentation, and keypoint estimation, have emerged as canonical tasks in visual recognition because of their intuitive appeal and pertinence in a wide variety of real-world problems.
The modus operandi followed in nearly all state-of-the-art visual recognition methods is the following: (i) Pre-train a large neural network on a very large and diverse image classification dataset, (ii) Append a small task-specific network to the pre-trained model and fine-tune the weights jointly on a much smaller dataset for the task. The introduction of ResNets by He \etal~\cite{he2016deep} made the training of very deep networks possible, helping in scaling up model capacity, both in terms of depth and width, and became a well-established instrument for improving the performance of deep learning models even with smaller datasets~\cite{kornblith2019better}. As a result, the past few years have seen increasingly large neural network architectures~\cite{mahajan2018exploring,zagoruyko2016wide,tan2019efficientnet,DBLP:journals/corr/abs-1811-06965}, with sizes often exceeding the memory limits of a single hardware accelerator.
In recent years, efforts towards reducing the memory and computation footprint of deep networks have followed three seemingly parallel tracks with common objectives: weight quantization, sparsity via regularization, and network pruning; Weight Quantization~\cite{hubara2017quantized,gong2014compressing,soudry2014expectation,cheng2015training,li2019fully} methods either replace weights of a trained neural network with lower precision or arithmetic operations with bit-wise operations to reduce the memory up to an order of magnitude. Regularization approaches, such as dropout~\cite{srivastava2014dropout,ba2013adaptive} or LASSO~\cite{tibshirani1996regression}, attempt to discourage an over-parameterized network from relying on a large number of features and encourage learning a sparse and robust predictor. Both quantization and regularization approaches are effective in reducing the number of weights in a network or the memory footprint, but usually at the cost of increased error rates~\cite{hastie2015statistical,li2019fully}. In comparison, pruning approaches~\cite{lecun1990optimal,han2015learning} disentangle the learning task from pruning by alternating between weight optimization and weight deletion. The recently proposed Lottery Ticket Hypothesis ·(LTH)~\cite{frankle2018the} falls in this category.

According to LTH, an over-parameterized network contains sparse sub-networks which not only match but sometimes even exceed the performance of the original network, all by virtue of a ``lucky'' random initialization before training. The original paper was followed up with tips and tricks to train large-scale models under the same paradigm~\cite{frankle2019lottery}. Since then, there has been a large, growing body of literature exploring its nuances.
Although some of these recent works have tried to answer the question -- how well do the tickets transfer across domains
~\cite{morcos2019one,mehta2019sparse}, when it comes to vision tasks -- the buck stops at image classification.

In this work, we aim to extend and explore the analysis of lottery tickets to fundamental visual recognition tasks of object detection, instance segmentation, and keypoint detection. Popular methods for such recognition tasks use a two-stage detection pipeline, with a supervised pre-trained convolutional neural network (ConvNet) backbone, a region proposal network (RPN), and one or more region-wise task-specific neural network branches. Loosely speaking, a ConvNet backbone is the most computationally intensive part of the architecture, and pre-training is the most time-consuming part. Therefore, as part of this study, we explore the following questions: (a) Are there \textit{universal} sub-networks within the ConvNet backbone that can be transferred to the downstream object recognition tasks? (b) Can we train sparser and more accurate sub-networks for each of the downstream tasks? And, (c) How does the behavior or properties of these sub-networks change with respect to the corresponding dense network? We investigate these questions under the dominant settings used in object recognition frameworks. Specifically, we use ImageNet~\cite{deng2009imagenet} pre-trained ResNet-18 and ResNet-50~\cite{he2016deep} backbones, Faster R-CNN~\cite{ren2015faster} and Mask R-CNN~\cite{he2017mask} modules for object recognition on Pascal VOC~\cite{everingham2010pascal} and COCO~\cite{lin2014microsoft} datasets. 
Our contributions are as follows:
\begin{itemize}
    \item 
    We show that tickets obtained from ImageNet training don't transfer to object recognition in case of COCO, \ie, there are no \textit{universal} tickets in pre-trained ImageNet models that can be used for downstream recognition tasks without a drop in performance. This is in contrast with previous works related to ticket transfer in vision models~\cite{mehta2019sparse,morcos2019one}. In case of smaller datasets such as Pascal VOC, we are able to find winning tickets from ImageNet pre-training with upto 40\% sparsity.
    \item With direct pruning, we can find ``task-specific'' tickets with up to 80\% sparsity for each of the datasets and backbones. We also investigate the efficacy of methods introduced by~\cite{frankle2018the,morcos2019one,frankle2019linear,renda2020comparing} such as iterative magnitude pruning, late resetting, early bird training, and layerwise pruning in the context of object recognition.
    \item Finally we analyse the behavior of tickets obtained for object recognition tasks, with respect to various task attributes such as object size, frequency, and difficulty of detection, to make some expected (and some surprising) observations. 
\end{itemize}

\section{Related Work}

\noindent\textbf{Model Compression:}
Ever since deep neural networks started gaining traction in real-world applications, there have been serious attempts made to reduce their parameters, intending to attain lower memory footprints~\cite{gong2014compressing,wu2016quantized,hubara2017quantized,soudry2014expectation,cheng2015training,li2019fully}, higher inference speeds~\cite{vanhoucke2011improving,denton2014exploiting,han2015deep} and potentially better generalization~\cite{arora2018stronger}. Amongst the various proposed techniques, model pruning approaches are predominant mainly due to their simplicity and effectiveness. One line of methods follow an unstructured process where insignificant weights are set to zero and are frozen for the rest of the training. The significance of weights are quantified either by magnitude ~\cite{han2015learning} or gradients during training time~\cite{lee2018snip}.
In structured pruning methods, relationships between pruned weights are taken into consideration, leading to pruning them in groups. Methods like ~\cite{wen2016learning} utilize Group Lasso regularization to prune redundant filter weights to enable structural sparsity, ~\cite{louizos2017learning} uses explicit $L0$ regularization to make weights within structures have exact zero values, and network slimming~\cite{liu2017learning} learns an efficient network by modelling the scaling factor of batch normalization layer.

\smallskip
\noindent\textbf{The Lottery Ticket Hypothesis:}
The introduction of Lottery Ticket Hypothesis by \cite{frankle2019lottery} opened a pandora's box of immense possibilities in the field of pruning and sparse models. The original paper was followed by \cite{frankle2019stabilizing} where the authors introduce the concept of "late resetting" which enabled the application of the hypothesis to larger and deeper models.~\cite{zhou2019deconstructing} followed up by proposing an extensive, in-depth analysis where they %
show that the resetting of the weights need not be to the exact initialization, but just need to the initial signs. 
\cite{frankle2019linear} probes the aspect of resetting further to show that the reason why LTH works is because of its ability to make the subnetwork stable to SGD noise. As far as theoretical guarantees are considered, ~\cite{malach2020proving} offers strong theoretical proofs for the experimental evidence of LTH.
\cite{grosse2020many} probed an orthogonal question about the number of possible tickets from a network. They showed that a single initialization had multiple winning tickets with low overlap and empirically conclude that there exists an entire ``distribution'' of winning lottery tickets. 

Complementary to LTH\cite{frankle2019lottery}, ~\cite{lee2018snip} and~\cite{wang2020picking} offer algorithms that can pick the winning ticket without the need for training. But they do not match the performance of the original procedure. The problem of longer training using LTH was effectively tackled by~\cite{you2019drawing} which introduced the concept of ``early bird tickets'' where the authors show that the winning tickets and their masks are obtained in the first few epochs of training, foregoing the need to train the original initialization till convergence. 
The intriguing properties of LTH led to a glut of works which investigated its eclectic aspects.~\cite{morcos2019one} scrutinize the generalization properties of winning tickets and offer empirical evidence that winning tickets can be transferred across datasets and optimizers, in the realm of image classification. The authors also discuss the learnt ``inductive biases'' of the tickets which may lead to worse performance of transferred tickets when compared with a ticket obtained from the same dataset.
\cite{mehta2019sparse} then proposed a variation of the theory titled ``transfer ticket hypothesis'' where they investigate the effectiveness of transferring a mask generated from source dataset to a target dataset.~\cite{desai2019evaluating} shows that the winning tickets do not perform simple overfitting to any domain and carry forward certain inherent biases which can prove useful in other domains too.
There have been many applications of LTH in the fields of NLP \cite{chen2020lottery} \cite{Movva2020DissectingLT} \cite{desai2019evaluating}\cite{Brix2020SuccessfullyAT} and Reinforcement Learning \cite{yu2020playing} \cite{you2019drawing} as well.

The work of \cite{kdmile} briefly analyzes LTH on single stage detectors such as YOLOv3~\cite{redmon2018yolov3} and achieves $90\%$ winning tickets, while maintaining the mAP on the Pascal VOC 2007 dataset. However, as they evaluate on light-weight and fast detectors, their mAP ($\sim56$) is much lower compared to networks like Faster R-CNN~\cite{ren2015faster} which reach mAP of $\sim69$ with just a ResNet-18 backbone. Their work is also limited to object detection and does not provide a detailed analysis of LTH for the task.
The idea for transferring subnetworks obtained from ImageNet to object detection tasks was concurrently discussed by \cite{chen2020selfsup}. 
For small datasets such as Pascal VOC, \cite{chen2020selfsup} observes that ImageNet tickets transfer for detection and segmentation tasks. However, we extend the analysis to the larger COCO dataset and show that this observation doesn't hold. We further build upon these results, to test out the generalization and transfer capabilities of winning lottery tickets across different object recognition datasets and tasks in computer vision.

\section{Background: Lottery Ticket Hypothesis}
LTH states that dense randomly-initialized neural networks contain sparse sub-networks which can be trained in isolation and can match the test accuracy of the original network. These sub-networks are called winning tickets and can be identified using an algorithm called Iterative Magnitude Pruning (IMP). Suppose the number of iterations for pruning is $T$ and we wish to prune $p\%$ of the network weights. The weights/parameters are represented by $w \in \mathbb{R}^n$ and the pruning mask by $m \in \{0,1\}^n$ where $n$ is the total number of weights in the network. The complete algorithm is presented in \ref{alg:IMP}.

\begin{algorithm}[t]
\caption{Iterative Pruning for LTH}\label{alg:IMP}
\begin{algorithmic}[1]
\State Randomly initialize network $f$ with initial weights $w_0$, mask $m_0 = \mathbbm{1}$, prune target percentage $p$, and $T$ pruning rounds to achieve it.
\While{$i < T$} %
\State Train network for N iterations $f(x; m_i \odot w_0) \rightarrow f(x; m_i \odot w_{i})$ 
\State Prune bottom $p^{\frac{1}{k}}\%$ of $m_i \odot w_{i} $ and update $m_i$.
\State Reset to initial weights $ w_0$ 
\State $i\gets i+1$ \Comment{next round}
\EndWhile\label{IMP}
\end{algorithmic}
\end{algorithm}

This pruning method can be one-shot when it proceeds for only a single iteration or it can proceed for multiple iterations, $k$, pruning $p^{\frac{1}{k}}\%$ each round. The authors also use other techniques such as learning rate warmup and show that finding winning tickets is sensitive to the learning rate.\\
While this method obtains winning tickets for smaller datasets, like MNIST \cite{lecun2010mnist}, CIFAR10 \cite{krizhevsky2009learning}, they fail to generalize to deeper networks, such as ResNets, and larger vision benchmarks, such as ImageNet \cite{deng2009imagenet}. \cite{frankle2019stabilizing} shows that IMP fails when resetting to the original initialization. They claim that resetting instead to the network weights after a few iterations of training provides greater stability and enables them to find winning tickets in these larger networks. They show that rewinding/late resetting to $3-7\%$ into training yields subnetworks which are $70\%$ smaller in the case of ResNet-50, without any drop in accuracy.

\begin{table*}
    \caption{Performance on the COCO dataset for ImageNet transferred tickets for ResNet-18 backbone at varying sparsity. The results for VOC are averaged over 5 runs with the standard deviation in parantheses. We obtain winning tickets at higher sparsity for smaller datasets like VOC compared to COCO.}
    \vspace{-0.1in}
    \begin{tabularx}{\linewidth} {D{0.8}*{9}{D{0.93}}D{1.0}D{1.6}}
    \toprule
    \multirow{2}{*}{Prune $\%$}&\multicolumn{3}{Y{3}}{\qquad COCO Detection}&\multicolumn{3}{Y{3}}{\quad \ \ COCO segmentation}&\multicolumn{3}{Y{3}}{\quad \ \ COCO Keypoint }& \multicolumn{2}{Y{2.6}}{\ \   VOC Detection }\\
    \cmidrule(lr){2-4}\cmidrule(lr){5-7}\cmidrule(lr){8-10}\cmidrule(lr){11-12}
    &Network sparsity&mAP&AP50&Network sparsity&mAP&AP50&Network sparsity&mAP&AP50&Network sparsity&mAP\\
    \cmidrule(lr){1-1}\cmidrule(lr){2-4}\cmidrule(lr){5-7}\cmidrule(lr){8-10}\cmidrule(lr){11-12}
    $90\%$& $31.61\%$& $25.59$& $43.69$& $31.61\%$& $24.03$& $40.89$& $21.47\%$& $55.30$& $79.30$& $79.49\%$& $63.91 (\pm0.41)$ \\
    $80\%$& $28.10\%$& $27.70$& $46.50$& $28.10\%$& $25.90$& $43.70$& $19.09\%$& $56.70$& $81.10$& $70.66\%$&$65.82 (\pm0.23)$\\
    $50\%$& $17.57\%$& $28.52$& $47.54$& $17.57\%$& $26.60$& $44.66$& $11.94\%$& $56.96$& $80.83$& $44.16\%$&$68.06(\pm0.11)$\\
    $0\%$& $0\%$& $29.91$& $49.05$& $0\%$& $27.64$& $46.00$& $0\%$& $58.59$& $82.04$& $0\%$&$68.53(\pm0.29)$\\
    \bottomrule
    \end{tabularx}
    
    \vspace{0pt}
    \label{tab:in_res18_coco}
\end{table*}
\begin{table*}
    \caption{Performance on the COCO dataset for ImageNet transferred tickets for  ResNet-50 backbone at various levels of pruning. The results for VOC are averaged over 5 runs with the standard deviation in parantheses. We obtain higher levels of sparsity compared to ResNet-18 transferred tickets which can be expected as it has fewer redundant parameters. Additionally, tickets for VOC have much higher sparsity with no drop in mAP compared to unpruned model.}
    \vspace{-0.1in}
    \begin{tabularx}{\linewidth} {D{0.8}*{9}{D{0.93}}D{1.0}D{1.6}}
    \toprule
    \multirow{2}{*}{Prune $\%$}&\multicolumn{3}{Y{3}}{\qquad COCO Detection}&\multicolumn{3}{Y{3}}{\quad \ \ COCO segmentation}&\multicolumn{3}{Y{3}}{\quad \ \ COCO Keypoint }& \multicolumn{2}{Y{2.6}}{\ \   VOC Detection }\\
    
    \cmidrule(lr){2-4}\cmidrule(lr){5-7}\cmidrule(lr){8-10}\cmidrule(lr){11-12}
    &Network sparsity&mAP&AP50&Network sparsity&mAP&AP50&Network sparsity&mAP&AP50&Network sparsity&mAP\\
    \cmidrule(lr){1-1}\cmidrule(lr){2-4}\cmidrule(lr){5-7}\cmidrule(lr){8-10}\cmidrule(lr){11-12}
    $90\%$& $41.99\%$& $30.66$& $50.75$& $41.99\%$& $28.68$& $47.76$& $31.49\%$& $57.78$& $82.25$&$65.37\%$& $71.20(\pm0.21)$\\
    $80\%$& $37.33\%$& $31.01$& $50.98$& $37.33\%$& $29.04$& $47.90$& $28\%$& $58.55$& $83.06$&$58.11\%$& $71.08(\pm0.20)$ \\
    $0\%$& $0\%$& $38.5$& $59.29$& $0\%$& $35.13$& $56.39$& $0\%$& $64.59$& $86.48$&0\%&$71.21(\pm0.32)$\\
    \bottomrule
    \end{tabularx}
    
    \vspace{-2pt}
    \label{tab:in_res50_coco}
\end{table*}

\section{LTH for Object Recognition}
In this section, we extend the Lottery Ticket Hypothesis to several object recognition tasks, such as Object Detection, Instance Segmentation, and Keypoint Detection. In \S\ref{ss:setup}, we describe the datasets, models, and metrics we use in our paper. \S\ref{ss:imagenet_transfer} examines the transfer of the lottery tickets obtained from ImageNet training to the downstream recognition tasks. 
\S\ref{ss:direct_pruning} investigates direct pruning on the downstream tasks.
\S\ref{ss:properties} analyzes the various properties of winning tickets obtained using direct pruning.

\subsection{Experimental setup}
\label{ss:setup}

We evaluate LTH primarily on the 2 datasets - Pascal VOC 2007 and COCO. We deal with the 3 tasks of object detection, instance segmentation, and keypoint detection for COCO and only object detection for VOC. 
We use Mask-RCNN for object detection and segmentation for COCO, Keypoint-RCNN for keypoint detection, and Faster-RCNN for object detection on VOC. We compare the results for ResNet-18 and ResNet-50 backbones. Note that while we use the term mean Average Precision (mAP) as a performance metric for all the tasks and datasets, the actual calculation of mAP is done using code provided in the respective datasets (and cannot be compared across the tasks).
\subsection{Transfer of ImageNet Tickets}
\label{ss:imagenet_transfer}
Many object recognition tasks utilize pre-trained networks whose backbones are trained on the ImageNet dataset. This is because ImageNet features and weights have shown the ability \cite{he2017mask} to generalize well to several downstream vision tasks. A plethora of works exists which perform LTH for the ImageNet classification task and obtain winning tickets. Therefore, tickets for standard convolutional backbones, such as ResNets, are readily available. This raises the pertinent question of whether pruned ImageNet trained models transfer directly to object recognition tasks. 
In order to answer this question, we transfer the pruned model to the backbone of the RCNN-based network and fine-tune the full network while ensuring the pruned weights in the backbone remain as zeros.

We perform experiments for the two architectures: ResNet-18 and ResNet-50, where we obtain $10\%,20\%$, and $50\%$ tickets on ImageNet by following the approach of \cite{frankle2019stabilizing}, and then transfer the model to the backbones of the three R-CNN based networks.
All models were trained on COCO, encompassing the three recognition tasks and the results are summarized in Tables \ref{tab:in_res18_coco}, \ref{tab:in_res50_coco}.
Additionally, we also perform similar experiments on the smaller Pascal VOC 2007 dataset for object detection to verify whether ImageNet tickets transfer without a significant drop in mAP. The results are shown in the last column of Tables \ref{tab:in_res18_coco}, \ref{tab:in_res50_coco}.
Note that pruning percentage of the ImageNet ticket is not equal to the actual network sparsity of the various networks as only the backbone of the networks are transferred and they make up a fraction of the total weights.

We see that ImageNet tickets transferred to COCO show a noticeable drop in mAP even with low levels of sparsity. We also note that another drawback of training transferred tickets on COCO is that they require careful tuning of learning rate and batch size for different tasks. On the other hand, for smaller datasets such as Pascal VOC, winning tickets are easily obtained at higher levels of sparsity which is $\sim45\%$ for ResNet-18 and $\sim65\%$ for ResNet-50. The larger networks can be pruned to a greater extent for both datasets.

ImageNet transferred tickets offer very little sparsity as the backbone of the networks usually do not make up most of the weights. For example, the ResNet-18 based Mask-RCNN for COCO Detection and Segmentation has only $~44\%$ of the parameters in the backbone, and hence, the overall network sparsity reaches $~31\%$ when pruning $90\%$ of the backbone weights, as shown in Table \ref{tab:in_res18_coco}. The rest of the weights are usually from the fully-connected layers of the network. It is therefore imperative to prune layers in addition to the backbone to increase network sparsity without decrease in mAP. As a consequence, we look into directly pruning the full network using LTH. %

\begin{figure*}[!ht]
\begin{tabularx}{\linewidth}{C{1.0}C{1.0}C{1.0}}
    \begin{subfigure}[]{\textwidth}
  \includegraphics[width=0.3\textwidth]{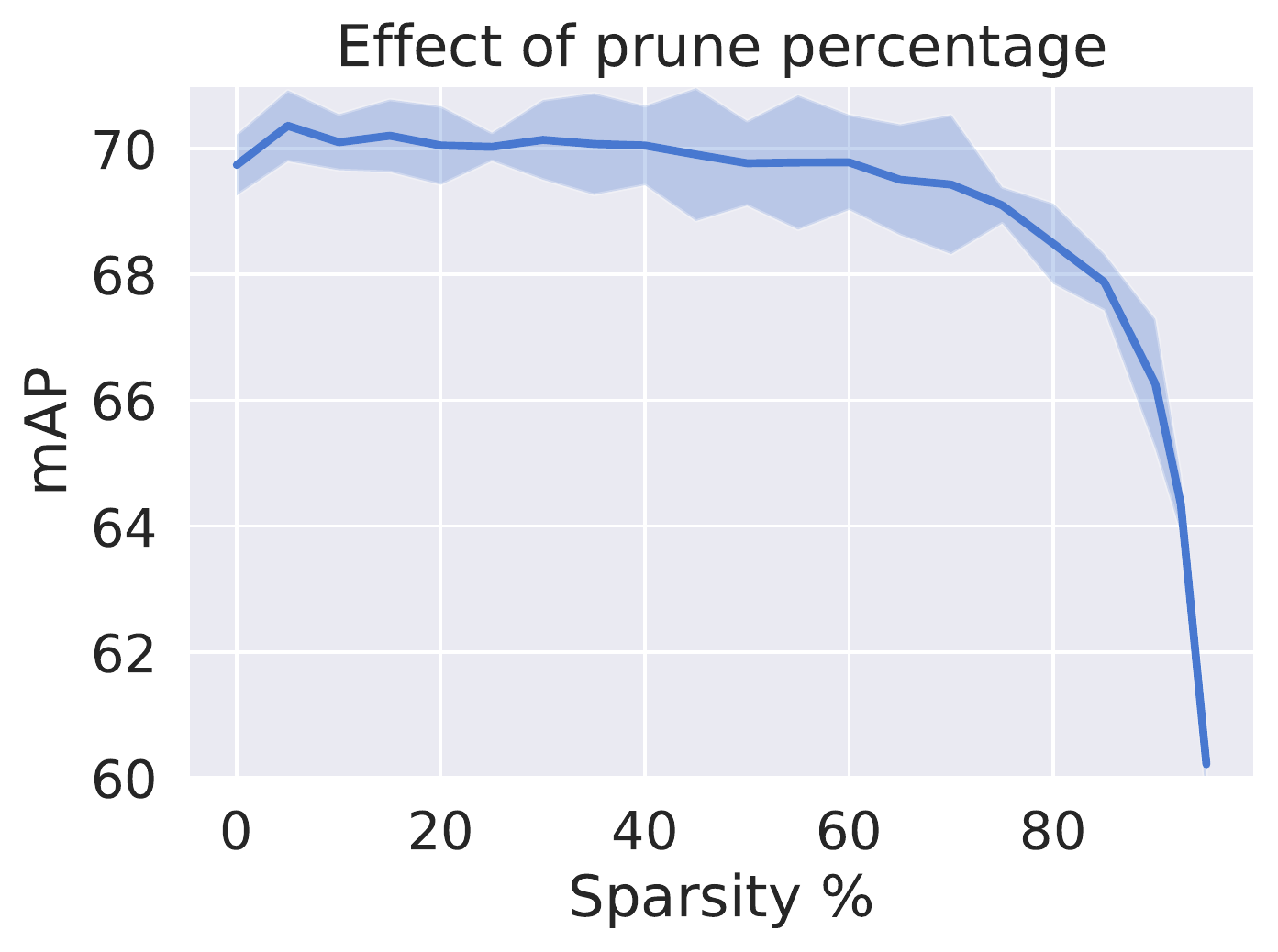}
\refstepcounter{subfigure}\label{fig:sparsity_voc}
    \end{subfigure}\hfil 
    & 
    \begin{subfigure}[]{\textwidth}
  \includegraphics[width=0.32\textwidth]{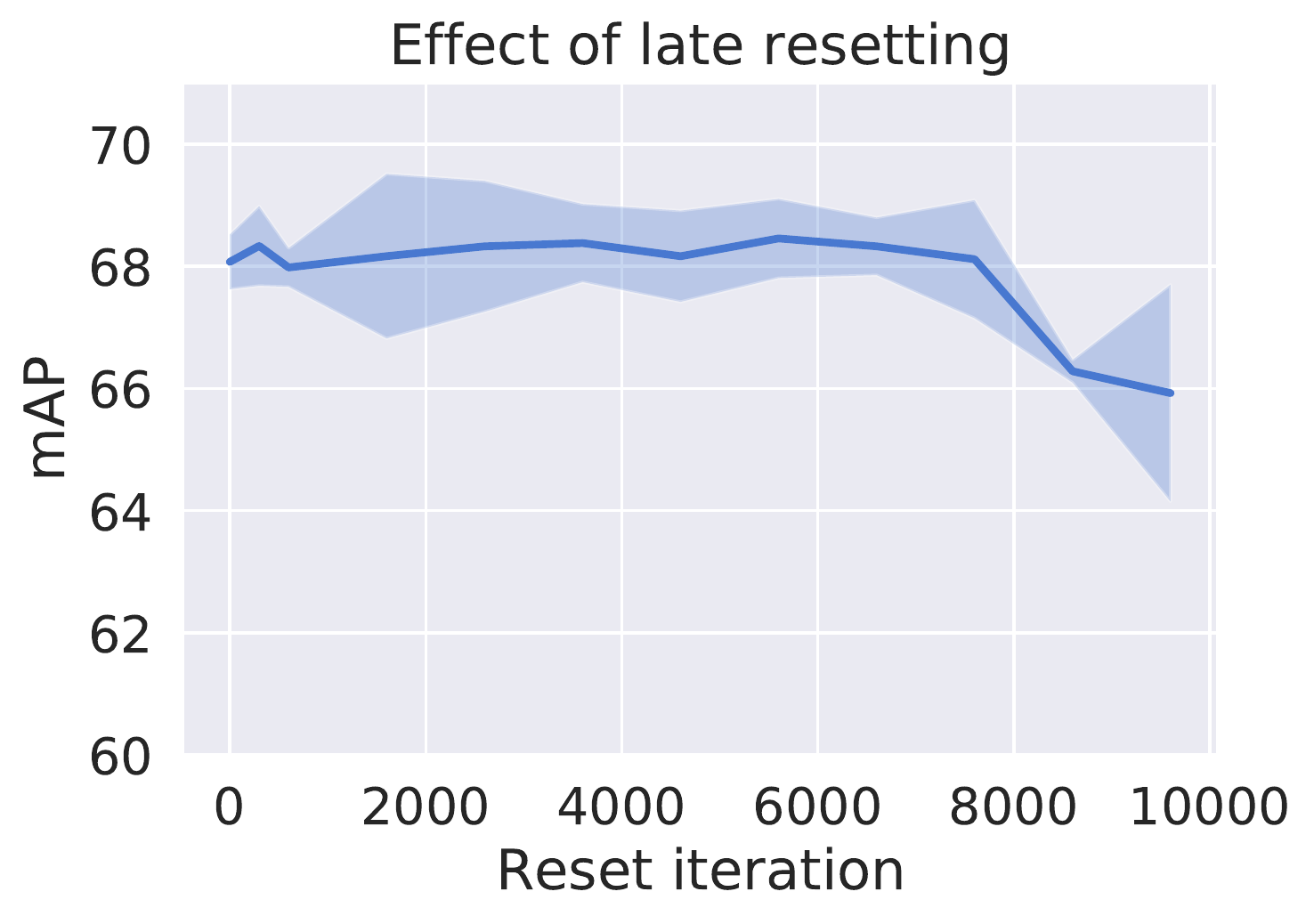}
\refstepcounter{subfigure}\label{fig:late_resetting}
    \end{subfigure}\hfil  
    &
    \begin{subfigure}[]{\textwidth}
  \includegraphics[width=0.3\textwidth]{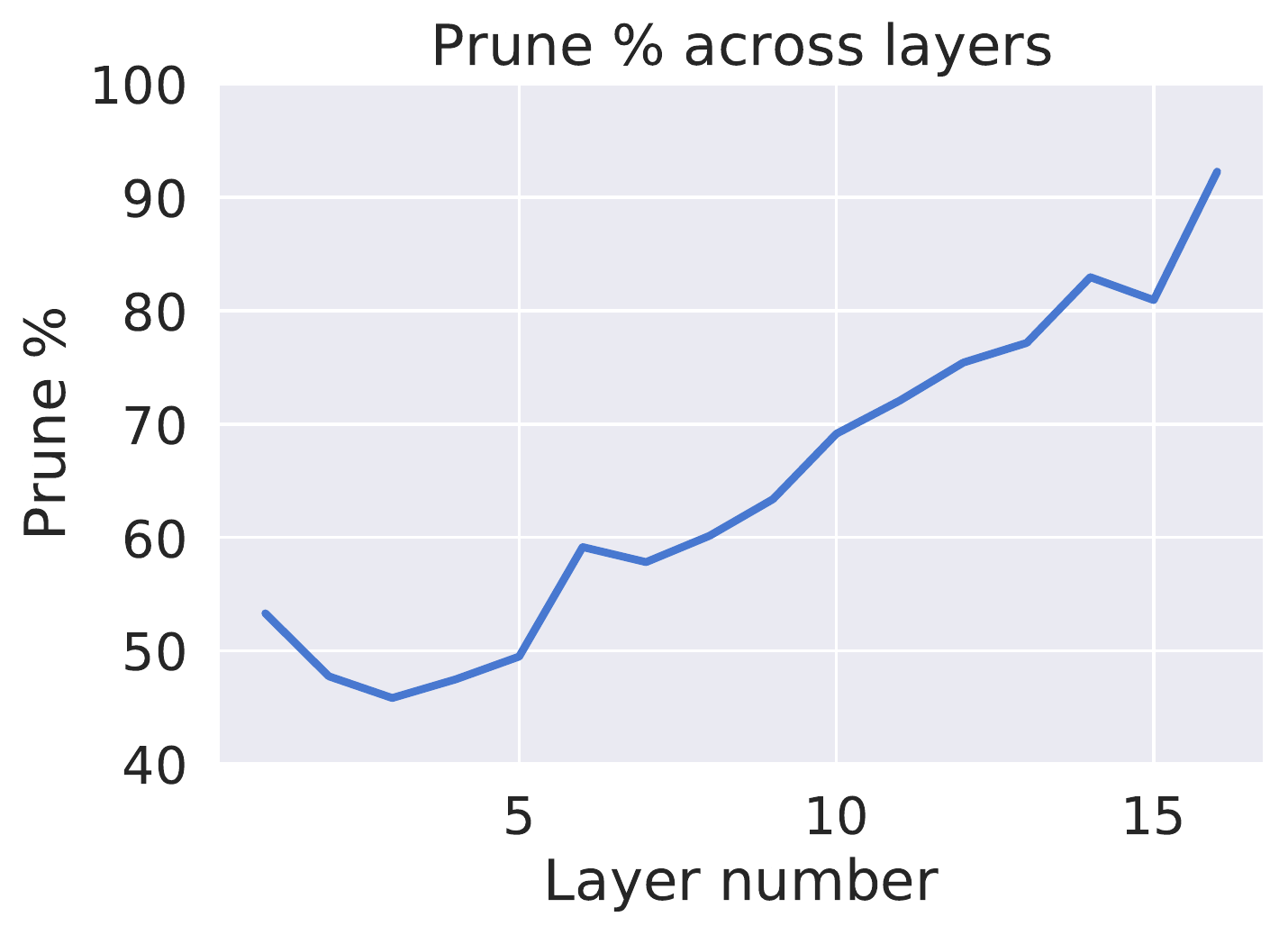}
\refstepcounter{subfigure}\label{fig:layer_pruning}
    \end{subfigure}\hfil  
 \\[-5pt]
\qquad\ \ (a)&\qquad \ \ (b)&\qquad \ \ (c)\\[5pt]
    \begin{subfigure}[]{\textwidth}
  \includegraphics[width=0.3\textwidth]{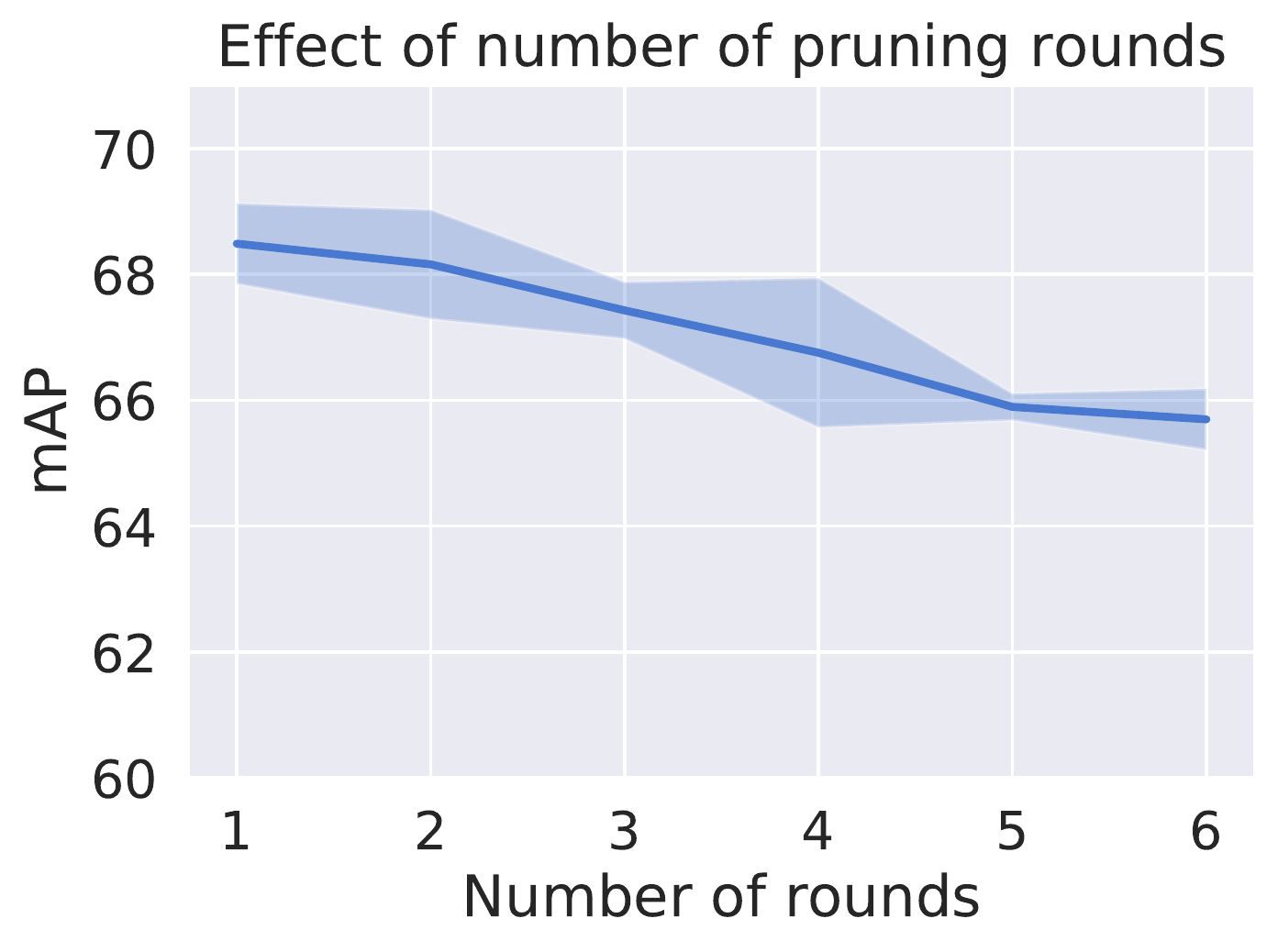}
\refstepcounter{subfigure}\label{fig:num_rounds}
    \end{subfigure}\hfil  
    &
    \begin{subfigure}[]{\textwidth}
  \includegraphics[width=0.3\textwidth]{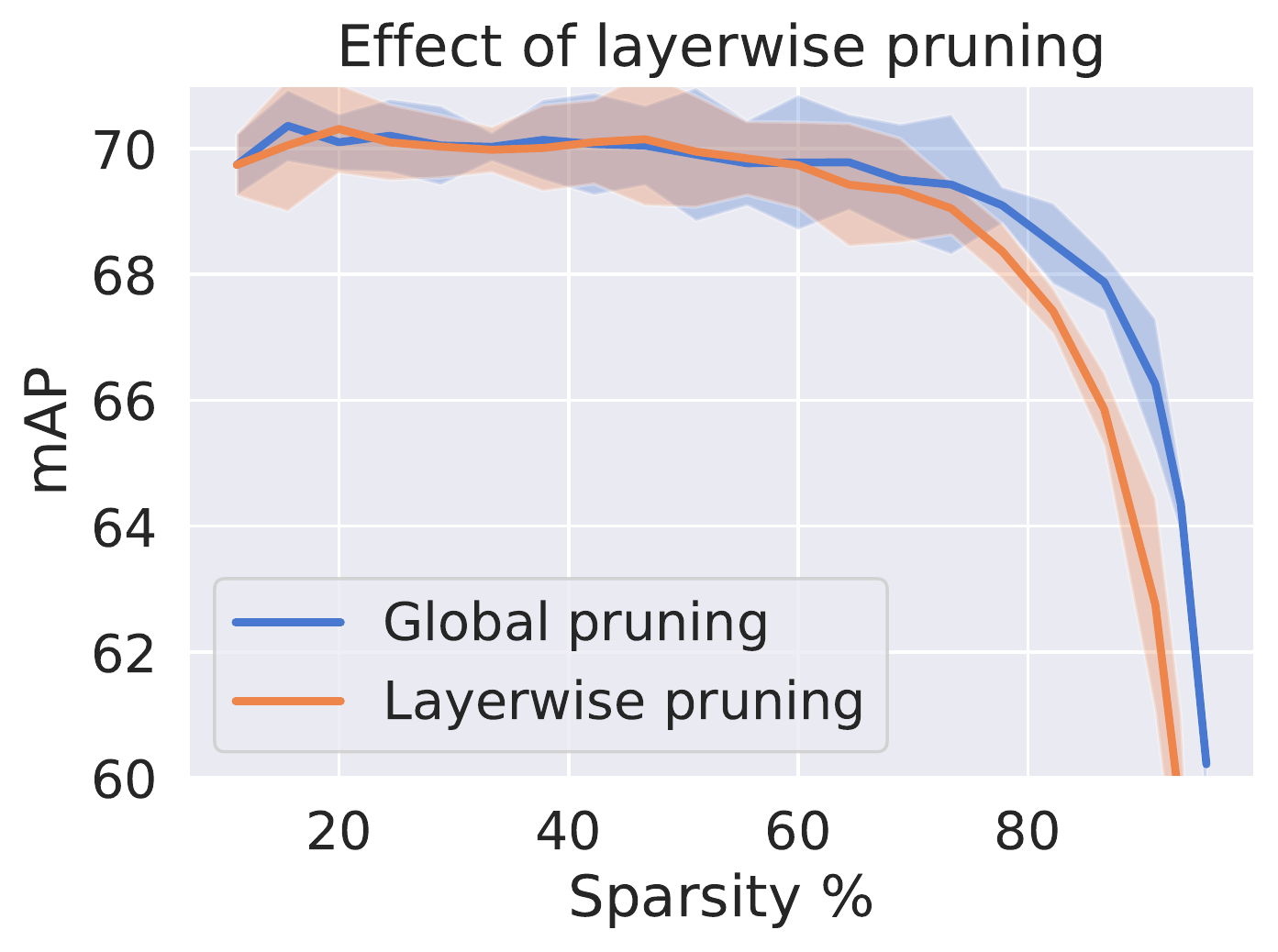}
\refstepcounter{subfigure}\label{fig:prune_type}
    \end{subfigure}\hfil  
    &
    \begin{subfigure}[]{\textwidth}
  \includegraphics[width=0.35\textwidth]{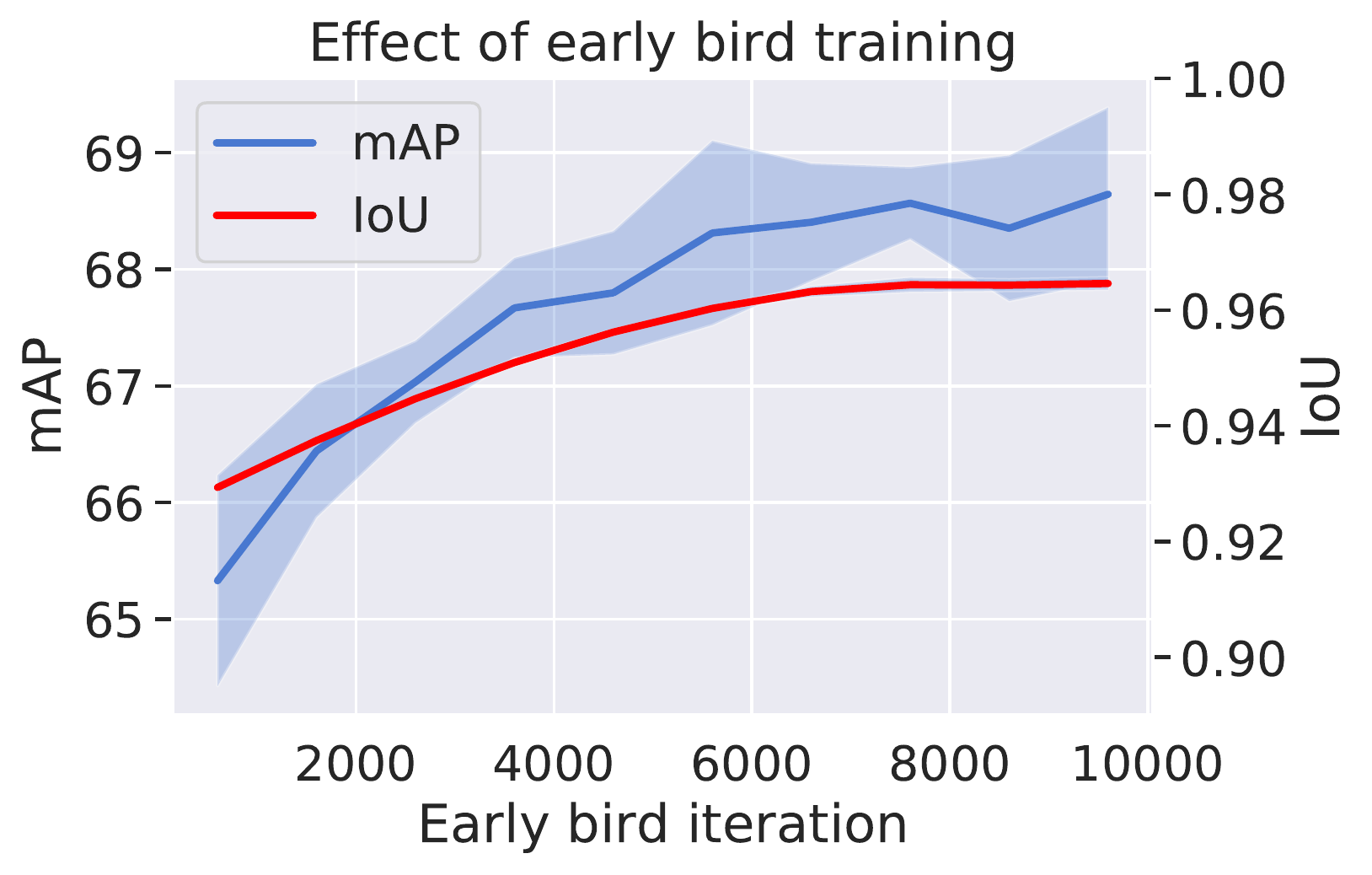}
\refstepcounter{subfigure}\label{fig:early_bird}
    \end{subfigure}\hfil 
 \\[-5pt]
\qquad\ \ (d)&\qquad \ \ (e)&\qquad \qquad \ \ (f)\\
\end{tabularx}
\caption{Effect of varying different hyperparameters for pruning Faster RCNN with ResNet-18 backbone on the Pascal VOC 2007~\cite{everingham2010pascal} dataset. All solid lines reported are the values averaged over 5 runs and the error bands are within $3$ times the standard deviation.}
\label{fig:hyperparams_voc}
\vspace{-12pt}
\end{figure*}

\subsection{Direct Pruning for Downstream Task}
\label{ss:direct_pruning}
In this section, we analyze the effect of various hyperparameters and pruning strategies for detection networks in order to obtain winning tickets. We primarily use the ResNet-18 backbone for Faster-RCNN trained on VOC for all our experiments in this section, unless mentioned otherwise. Even though the ResNet-18 backbone is smaller than other backbone networks such as ResNet-50, we find that similar conclusions hold for the larger networks as well.

The Faster RCNN network consists of parameters which we group into 4 main modules: Base Convolutions, Classification Network Convolutions  (Top), Region Proposal Network (RPN), Classification network box and classification fully connected heads (Box and Cls Head). We provide a detailed analysis of pruning these 4 groups and their effect on winning tickets. Additionally, we also analyze different pruning strategies and the role played by hyperparameters. %

\smallskip
\noindent\textbf{Varying Pruning Percentage:} We evaluate the network performance at varying levels of sparsity. We prune different percentages of parameters in the Base and Top modules which include $~88\%$ of the total network parameters. The results are plotted in Fig.\ref{fig:hyperparams_voc}\subref{fig:sparsity_voc}. We achieve performance within one standard deviation of the baseline, with $70\%$ sparsity. Our models outperform the baseline mean, thereby proving that we can indeed obtain high performance winning tickets at much higher levels of sparsity for detection. Additionally, we see that at any given sparsity, direct pruning yields much better results compared to ImageNet transferred tickets.
We also show that these observations pan other vision tasks by obtaining winning tickets for Mask-RCNN and Keypoint-RCNN with both ResNet-18 and ResNet-50 backbones on the COCO dataset. We additionally prune FC layers in these set of experiments in order to achieve desired sparsity levels as they take up $\sim50\%$ of the total weights. The results for ResNet-18 are shown in Fig.~\ref{fig:teaser}. We obtain winning tickets with $80\%$ sparsity on all the three tasks while outperforming the unpruned network for lower levels of sparsity. Additionally, we consistently outperform the different ImageNet transferred tickets ($50\%$, $80\%$, $90\%$) by a large margin supporting our claim that direct training of tickets on downstream tasks yield better results than ImageNet tickets.

\smallskip
\noindent\textbf{Effect of Early/Late Resetting:} \cite{frankle2019stabilizing} states that resetting the network to a few iterations through training instead of the initialization stabilizes the winning ticket training. We evaluate whether this holds true for detection tasks as well.
We show the performance of winning tickets as a function of resetting at various stages of training in Fig.~\ref{fig:hyperparams_voc}\subref{fig:late_resetting} and observe that resetting during the earlier or even mid stages of training does not have a very strong effect on the final mAP. This is likely because the backbones of detection networks are initialized with ImageNet weights and are not random as is the case with other papers dealing with LTH in the classification setting. Therefore, the weights are more stable and late resetting is not necessary.
We also additionally analyze effects of resetting towards the end of training and notice that there is a sharp drop in the performance after $8k$ iterations. This is because the learning rate is decayed at this stage of training and the parameters change significantly right after. A similar case holds when we perform learning rate warmup but do late resetting before the learning rate is fully warmed up. The performance drops significantly as the learning rate keeps fluctuating showing that late resetting is quite sensitive to learning rate.

\smallskip
\noindent\textbf{Pruning different Faster-RCNN modules:} We prune $20\%$ of the parameters of the various modules within the Faster-RCNN network and analyze their effects on the mAP. 
We also try different combinations of pruning with the modules and report the results in Table \ref{tab:modules}. Pruning the Box and Classification head (which takes up only ~$65\%$ weights) outperforms the baseline case of no pruning, but does not always improve performance when other modules are being pruned. Additionally, pruning the $RPN$ module increases the performance slightly even though it comprises of only $~10\%$ of the network weights. Next, pruning the Base module and/or the Top module of the backbone leads to a drop in performance, which is expected as they consist of $~22\%$ and $~66\%$ of the weights respectively. Pruning the Base alone, excluding the Top, performs nearly as well as the baseline, while including the Top yields a lower mAP.

\begin{table}\footnotesize\centering
\renewcommand{\arraystretch}{1.2}
\renewcommand{\tabcolsep}{6pt}
    \caption{Performance on Pascal VOC by pruning different modules of a ResNet-18 Faster-RCNN network. The results are averaged over 5 runs with the standard deviation in parantheses. $\checkmark$ represents the module being pruned, while Param $\%$ represents the percentage of parameters occupied by the modules being pruned.}
    \vspace{-0.1in}
    \resizebox{\linewidth}{!}{
    \begin{tabular}{@{}ccccccc@{}}
    \toprule
    Base &Top &RPN & \makecell{Box, \\Cls Head}&\makecell{Param\\ $\%$}& \makecell{Network\\ Sparsity}&mAP\\\midrule
    -&-&-&-&$0$&$0\%$&$69.74$ ($\pm$0.16)\\
    -&-&-&\checkmark&$0.65$&$0.52\%$&$70.30$ ($\pm0.14$)\\
    -&-&\checkmark&-&$9.71$&$7.77\%$&$70.02$ ($\pm0.19$)\\
    -&-&\checkmark&\checkmark&$10.36$&$8.29\%$&$70.08$ ($\pm0.10$)\\
    \checkmark&-&-&-&$21.93$&$17.55\%$&$69.32$ ($\pm0.07$)\\
    \checkmark&-&-&\checkmark&$22.59$&$18.07\%$&$69.60$ ($\pm0.19$)\\
    \checkmark&-&\checkmark&-&$31.64$&$25.31\%$&$69.39$ ($\pm0.25$)\\
    \checkmark&-&\checkmark&\checkmark&$32.29$&$25.83\%$&$69.47$ ($\pm0.15$)\\
    -&\checkmark&-&-&$66.39$&$53.11\%$&$69.02$ ($\pm0.19$)\\
    -&\checkmark&-&\checkmark&$67.04$&$53.63\%$&$68.74$ ($\pm0.21$)\\
    -&\checkmark&\checkmark&-&$76.09$&$60.88\%$&$68.88$ ($\pm0.25$)\\
    -&\checkmark&\checkmark&\checkmark&$76.75$&$61.40\%$&$68.93$ ($\pm0.26$)\\
    \checkmark&\checkmark&-&-&$88.32$&$70.66\%$&$68.45$ ($\pm0.21$)\\
    \checkmark&\checkmark&-&\checkmark&$88.97$&$71.18\%$&$68.54$ ($\pm0.23$)\\
    \checkmark&\checkmark&\checkmark&-&$98.03$&$78.42\%$&$68.51$ ($\pm0.23$)\\
    \checkmark&\checkmark&\checkmark&\checkmark&$98.68$&$78.94\%$&$68.47$ ($\pm0.10$)\\
    \bottomrule
    \end{tabular}
    }
    \label{tab:modules}
    \vspace{-.2in}
\end{table}

\smallskip
\noindent\textbf{Performance of Early-bird tickets:} ~\cite{you2019drawing} showed that tickets can be found at early stages of training. We visualize this by obtaining masks at various stages in training and evaluating their performance. We also plot each masks' Intersection over Union (IoU) with the default mask obtained at the end of training. This IoU shows the overlap in the parameters being pruned. The results are visualized in Fig.~\ref{fig:hyperparams_voc}\subref{fig:early_bird}. We see that within $50\%$ of network training we find tickets whose performance is within a standard deviation of the performance of the default ticket (obtained at the end of training). This is because the IoU becomes more or less stable at around $~0.96$ during the middle stages of training and the mask is unchanged as training advances. This allows us to cut down on the number of training iterations significantly with very little cost to the network performance.
\begin{figure*}
    \centering
    \includegraphics[width=\textwidth]{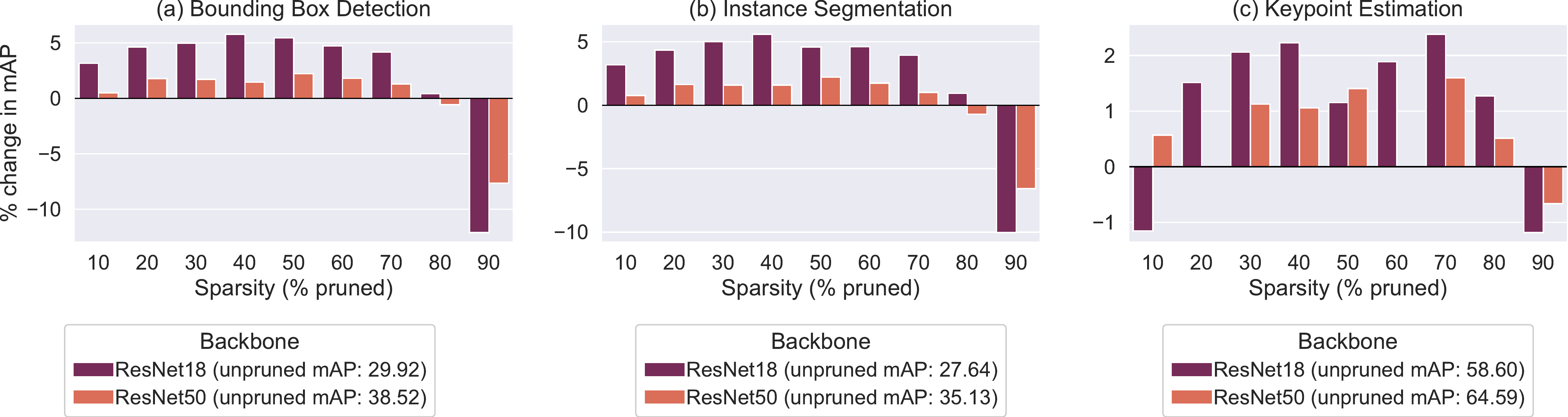}
    \caption{ResNet-18 \vs ResNet-50. We analyse change in mAP by using LTH on Mask R-CNN with different backbones.}
    \label{fig:architecture}    
\end{figure*}

\smallskip
\noindent\textbf{Effect of number of rounds of pruning:} \cite{frankle2018the} states that iterative pruning performs better than one-shot pruning on the classification task with small datasets and networks. We show that this does not necessarily hold true for detection and larger backbones. We plot the network's performance against various rounds of pruning and observe that one-shot pruning outperforms iterative methods in Fig.~\ref{fig:num_rounds}.

\smallskip
\noindent\textbf{Layer-wise \vs global pruning:} \cite{frankle2019stabilizing} performs global pruning for larger datasets and networks and claims that pruning at the same rate in lower layers as compared to higher layers, is detrimental to the network performance. We evaluate the two methods of pruning on the detection task and show the results in Fig.~\ref{fig:hyperparams_voc}\subref{fig:prune_type}. Additionally, for global pruning, we plot the percentage of parameters pruned in each layer of the backbone network in Fig.~\ref{fig:layer_pruning}. Layer-wise pruning does as good as global pruning for lower levels of sparsity. However, there is a noticeable performance gap for sparsity levels above $60\%$.
This is because layer-wise pruning forces lower layers with very few parameters to have high sparsity percentages. But as per Fig.~\ref{fig:layer_pruning}, for global pruning, we see that lower layers are pruned less as they are crucial to both the RPN and Classification stages of the network.

\subsection{Properties of Winning Tickets}
\label{ss:properties}

In Section~\ref{ss:direct_pruning}, we showed that we can discover sparser networks within our two-stage Mask-RCNN detector if we directly prune on the task itself. We build upon those results to further probe the properties of winning tickets.

\smallskip
\noindent\textbf{Effect of backbone architecture:}
In Fig.~\ref{fig:architecture}, we show how winning tickets behave for 2 different backbones, ResNet-18 and ResNet-50, at different sparsity levels (50\%, 80\%, 90\%).
We make two observations: First, the breaking point for both networks is $\sim80\%$ sparsity. However, performance of ResNet-18 drops more sharply than ResNet-50 afterwards. This is intuitive since ResNet-18 has fewer redundant parameters and over-pruning leads to drop in the performance. Second, as we gradually increase the sparsity of the networks, mAP increases for all tasks in case of both networks. However, gains for ResNet-18 models are consistently more than ResNet-50.

\begin{figure*}[!ht]
\centering
    \includegraphics[width=\linewidth]{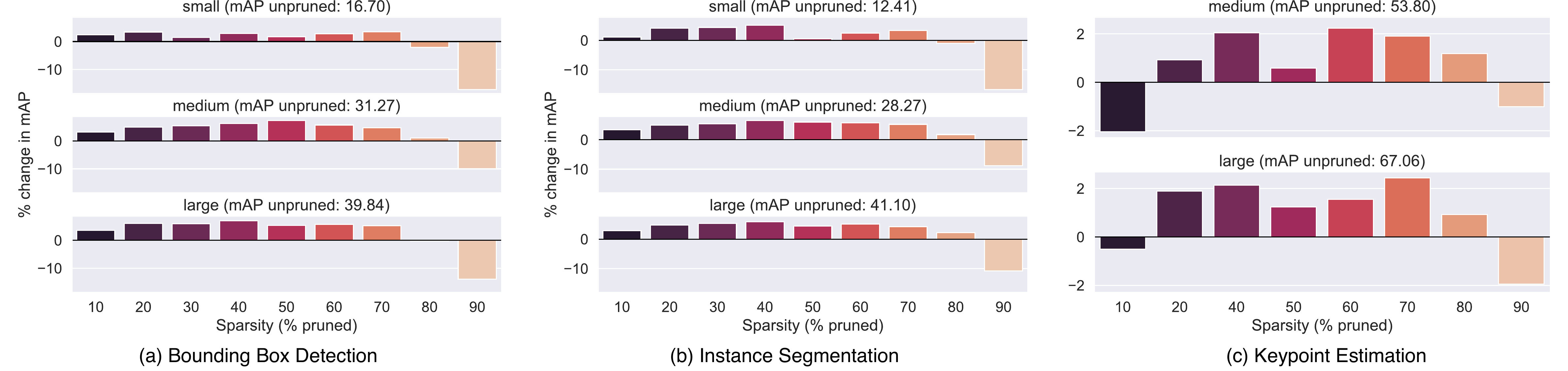}
    \vspace{-0.2in}
    \caption{Comparison of Mean Average Precision (mAP) of pruned model for different object sizes in case of Object Detection, Instance Segmentation, Keypoint Estimation. x-axis shows the sparsity of the subnetwork (or the percentage of weights removed). y-axis shows the percentage drop in mAP as compared to the unpruned network. For all tasks, and object sizes, performance doesn't drop till about 80\% sparsity. After which, small objects are hit slightly harder as compared to medium and large objects.}
    \label{fig:boxsize_pruning}
    \vspace{-.1in}
\end{figure*}

\smallskip
\noindent\textbf{Do winning tickets behave differently for varying object sizes?}
Using the definition from ~\cite{lin2014microsoft}, we categorize bounding boxes into small (area $<32^2$), medium ($32^2<$ area $<96^2$), and large (area $>32^2$). To understand how sparse networks behave for different sized objects, we plot the percentage gain or drop from the mAP of a dense network. Figure~\ref{fig:boxsize_pruning} shows the percentage change in mAP for different levels of sparsity in the Mask R-CNN model. We can observe that in each case, the model performance increases with sparsity, until sparsity reaches 80\%, after which, mAP sharply declines. We note that the percentage drop for small boxes is more, with winning tickets (10\% of weights) showing a drop of over 17\% in case of detection and segmentation tasks while medium sized objects show smaller drops than large objects for all tasks.

\begin{figure*}[!ht]
\centering
    \includegraphics[width=0.99\linewidth]{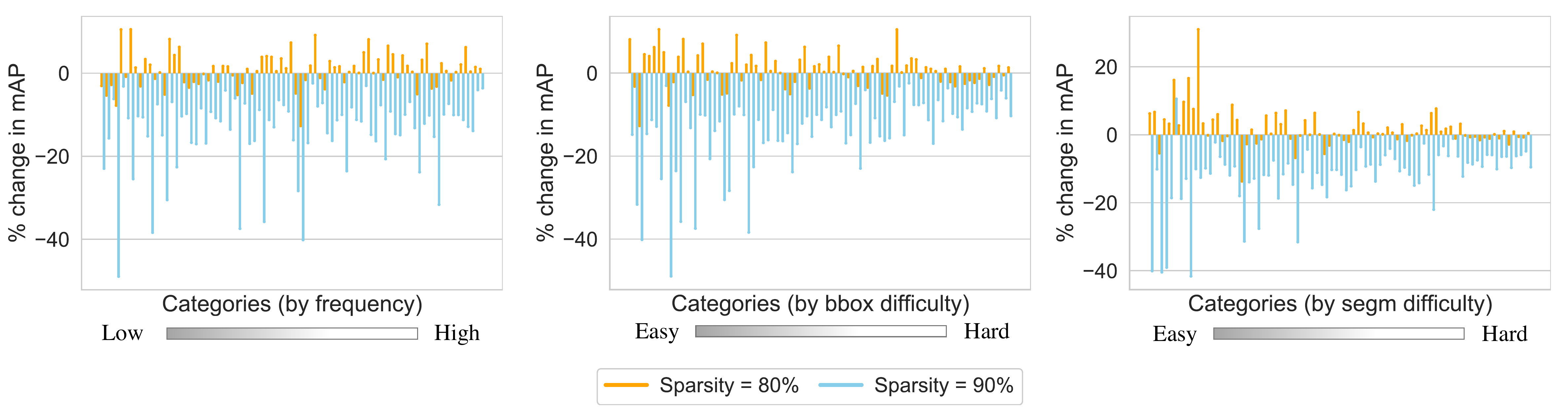}
    \vspace{-1em}
    \caption{Comparison of Mean Average Precision (mAP) of pruned model for 80 COCO object categories. x-axis in each of the plot is a list of categories (sorted using different criteria). y-axis shows the percentage drop in mAP as compared to the unpruned network.}
    \label{fig:cat_pruning}
    \vspace{-1em}
\end{figure*}

\smallskip
\noindent\textbf{How does the performance of the pruned network vary for rare \vs frequent categories?}
We sort the 80 object categories in COCO by their frequency of occurrence in training data. We consider networks with 80\% and 90\% of their weights pruned and observe the percentage change in the bounding box mAP of the model with respect to the unpruned network for each of the categories. Figure~\ref{fig:cat_pruning}(a) depicts the behavior with a bar graph.
While for most categories, winning tickets are obtained at 80\% sparsity, performance drops sharply with more pruning in case of rare categories (such as toaster, parking meter, and bear) as compared to common categories (such as person, car, and chair).

\smallskip
\noindent\textbf{Do the winning tickets behave differently on easy vs hard categories?}
For a machine learning model, an object can be easy or hard to recognize because of a variety of reasons. We have already discussed two reasons that influence the performance --- number of instances available in the training data, and size of the object. There can also be other causes that can render an object unrecognizable in given surroundings. Camouflage or occlusion, poor camera quality, light conditions, distance from the camera, or just variations within different instances or views of the object are few of them. Since exhaustive analyses of these causes is intractable, we rank object categories based on performance of an unpruned Mask R-CNN model. We do this categorization for detection and segmentation models as shown in Figure~\ref{fig:cat_pruning}(b) and (c). Note that `easy' and `hard' categories from these two definitions have an overlap but they are not the same. For example, knife, handbag, and spoon are the categories with lowest bounding box mAP, and giraffe, zebra, and stop signs are one with the highest (excluding `hair drier' which has 0 mAP). On the other hand, skis, knife, and spoon have the lowest segmentation mAP, while stop sign, bear, and fire hydrant have the highest. From the Figure~\ref{fig:cat_pruning}(b) and (c), we make the following observations --- (i) tickets with 80\% sparsity can actually increase mAP for certain categories like snowboard by as much as 38\%, (ii) Going from 80\% to 90\% sparsity, mAP drops significantly for easy categories compared to hard categories, (iii) categories that are hit the hardest such as skis, hot dog, spoon, fork, handbags usually have long, thin appearance in images.

\smallskip
\noindent
\textbf{Do winning tickets transfer across tasks?} We showed that ImageNet tickets transfer to a limited extent to downstream tasks. We further study whether the tickets obtained from the downstream task of detection/segmentation transfer to keypoint estimation and vice-versa. We train Mask-RCNN and Keypoint-RCNN respectively for the two tasks on the COCO dataset while maintaining a sparsity level of 80\%. For both the tasks we transfer all values till box head modules, after which the model structures differ. The results are shown in Table \ref{tab:task_transfer}. We can observe that the drop is marginal for the transfer of tickets between detection-segmentation to keypoint task, as compared with the reverse case which registers a significant drop. This might be because the ticket is obtained on the keypoint task which is trained only on `human' class and it fails to transfer well for the detection task which uses the entire COCO dataset. 

\begin{table}
    \caption{Effect of ticket transfer across tasks. Transferred tickets do worse than direct training as expected, but still do not result in drastic drops in the mAP or AP50. Here we do task transfer using the 80\% pruned model. }
    \begin{tabularx}{\linewidth} {D{0.9}D{1.1}D{1.2}D{0.8}D{0.8}}
    \toprule
    Target task&Source task&Network sparsity&mAP&AP50\\
    \midrule
    \multirow{2}{*}{Det}& Det/Seg&  78.4\%& 30.04& 49.40\\
    & Keypoint& 50.11\%& 23.94& 41.08\\\hdashline
    \multirow{2}{*}{Seg}& Det/Seg&78.4\%& 27.90& 46.68\\
    & Keypoint& 50.11\%& 23.02& 39.01\\\hdashline
    \multirow{2}{*}{Keypoint}& Det/Seg&  76.98\%& 58.31& 81.53\\
    & Keypoint& 79.4\%& 59.34& 82.36\\
    \bottomrule
    \end{tabularx}
    \label{tab:task_transfer}
    \vspace{-0.2in}
\end{table}

\section{Discussion}

\cite{mehta2019sparse,morcos2019one} show that winning tickets transfer well across datasets. However, the study in~\cite{mehta2019sparse} was limited to smaller datasets, like CIFAR-10 and FashionMNIST, and both~\cite{mehta2019sparse,morcos2019one} are limited to classification tasks. We obtain contrasting results when transferring tickets across tasks as shown in Sec.~\ref{ss:imagenet_transfer}. ImageNet tickets transfer with approximately $~40\%$ sparsity to fall within one standard deviation of the baseline network. This is likely due to the fact that winning tickets retain inductive biases from the source dataset which are less likely to transfer to a new domain and task. 
Additionally, we show that unlike prior LTH works, iterative pruning degrades the performance of subnetworks on detection and one-shot pruning provides the best networks. We also observe that due to the use of pre-trained weights from ImageNet for the backbone of detection networks, late resetting is not necessary for finding winning tickets. This is in contrast to the ~\cite{frankle2019stabilizing}, which is restricted to the classification task involving random initialization for the networks.
Like previous works, in our experiments as well, we find that sparse lottery tickets often outperform the dense networks themselves. However, we make another interesting observation --- in each of object recognition tasks, tickets with fewer parameters such as ResNet-18 show more gains in performance as compared to tickets with more parameters (ResNet-50).
We also find that small and infrequent objects face higher performance drop as the sparsity increases.

\section{Conclusion}
We investigate the Lottery Ticket Hypothesis in the context of various object recognition tasks. Our study reveals that the main points of original LTH hold for different recognition tasks, \ie, we can find subnetworks or winning tickets in object recognition pipelines with up to 80\% sparsity, without any drop in performance on the task. These tickets are task-specific, and pre-trained ImageNet model tickets don't perform as well on the downstream recognition tasks. We also analyse claims made in recent literature regarding training and transfer of winning tickets from an object recognition perspective. Finally, we analyse how the behavior of sparse tickets differ from their dense counterparts.
In the future, we would like to investigate how much speed up can be achieved using these sparse models with various hardware~\cite{lu2019efficient} and software modifications~\cite{elsen2020fast}. Extending this analyses for even bigger datasets such as JFT-300M~\cite{300m_iccv17} or IG-1B~\cite{mahajan2018exploring} and for self-supervised learning techniques is another direction to pursue.

\medskip
\noindent \textbf{Acknowledgements.} \small This work was partially supported by DARPA GARD \#HR00112020007 and a gift from Facebook AI.

{\small
\bibliographystyle{ieee_fullname}
\bibliography{egbib}
}
\clearpage

\appendix
\appendixpage
\maketitle

We provide additional details for some of the experiments presented in the paper. In particular, we provide comparison with a simpler ImageNet ticket transfer alternative in Section~\ref{sec:mask_transfer}, compare the different errors made by dense and pruned models in Section~\ref{sec:errors}, and finally verify the faster convergence of sparser models in Section~\ref{sec:convergence}.

\section{Mask Transfer Without Retraining}
\label{sec:mask_transfer}
\begin{figure}[!ht]
\centering
\includegraphics[width=\linewidth]{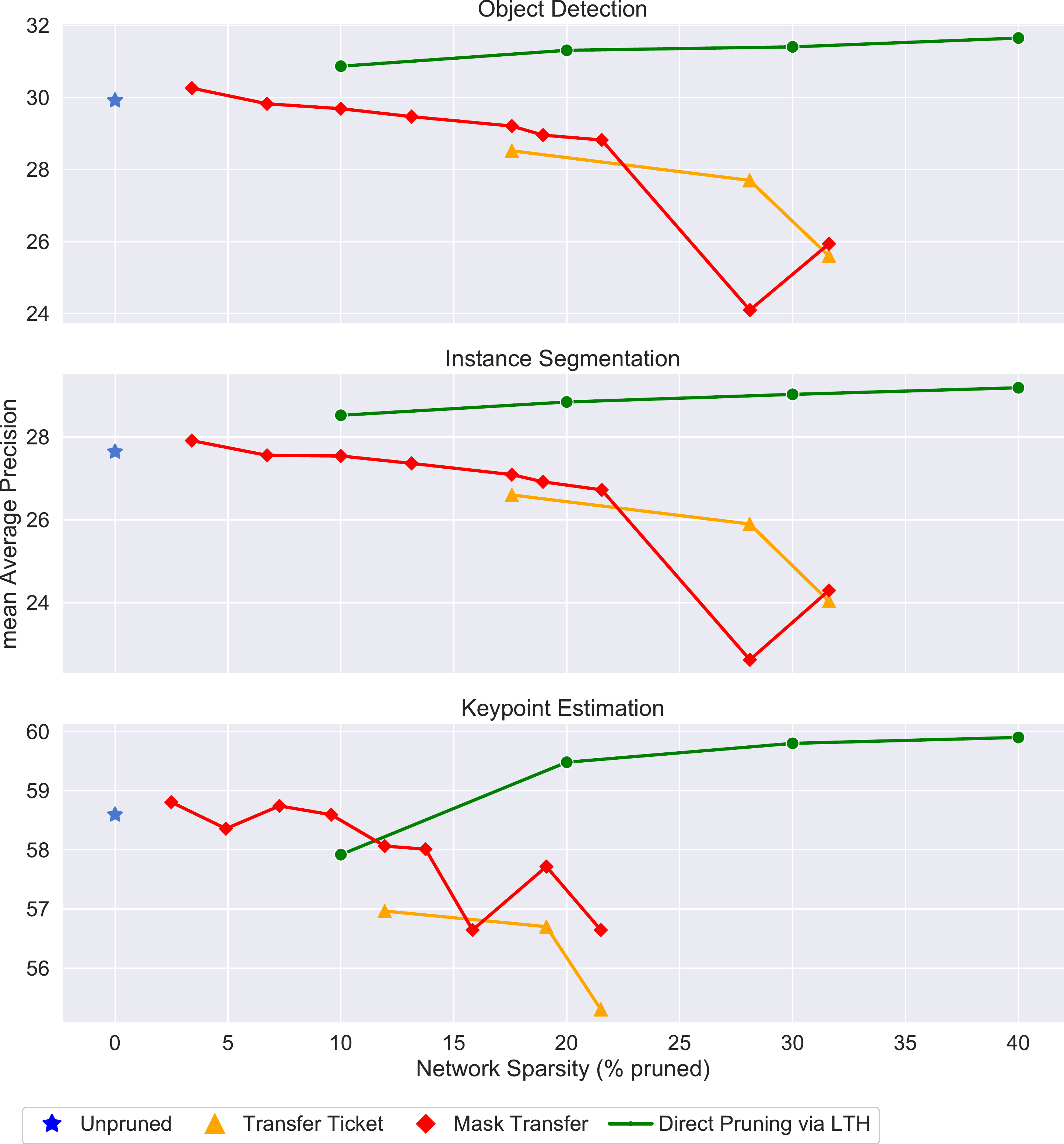}
\caption{Transferring ImageNet backbone tickets to object recognition tasks \vs Direct pruning via LTH on the object recognition tasks. We experiment with two variations of transferring ImageNet backbone tickets to object recognition tasks. `Transfer ticket' refers to the case when we transfer the lottery ticket backbone trained on ImageNet data to downstream task (also discussed in the Section 4 of the paper). `Mask Transfer' refers to the case when ticket is transferred without retraining on ImageNet, \ie, only the relevant mask from backbone is transferred keeping ImageNet weights the same. Best viewed in color.}
\label{fig:mask_transfer}
\end{figure}

In Section 4.2, we analyzed the effects of transferring tickets only for the ImageNet trained backbones. While this deals with transferring the ticket mask as well as values, we further analyze whether transferring only the mask provides winning tickets for these tasks using the methodology from \cite{mehta2019sparse}. 
We use the default ImageNet weights in the ResNet-18 and ResNet-50 backbone and keep the top $p\%$
of the weights in convolutional layers while setting the rest to zeros and maintaining it throughout the training of the entire network.
We refer to this method as `Mask Transfer'. Since training the backbone on much larger ImageNet data is performed only once, `Mask Transfer' is a much cheaper or computationally efficient way of obtaining tickets from parent task. We observe that behavior of `Mask Transfer` is similar to the `Transfer Ticket` obtained by method discussed in Section 4.2 where the sparse subnetwork weights are fully retrained on ImageNet. Either cases are outperformed by direct pruning on the downstream tasks. The results are summarized in 
Figure~\ref{fig:mask_transfer} (ResNet-18) and Table \ref{tab:in_res50_coco_mask} (ResNet-50).

\begin{table*}[!ht]
    \caption{Performance on the COCO dataset for ImageNet backbones with mask transfer tickets for  ResNet-50 at various levels of pruning. The results for VOC are averaged over 5 runs with the standard deviation in parantheses.}

    \begin{tabularx}{\linewidth} {D{0.8}*{9}{D{0.93}}D{1.0}D{1.6}}
    \toprule
    \multirow{2}{*}{Prune $\%$}&\multicolumn{3}{Y{3}}{\qquad COCO Detection}&\multicolumn{3}{Y{3}}{\quad \ \ COCO segmentation}&\multicolumn{3}{Y{3}}{\quad \ \ COCO Keypoint }& \multicolumn{2}{Y{2.6}}{\ \   VOC Detection }\\
    
    \cmidrule(lr){2-4}\cmidrule(lr){5-7}\cmidrule(lr){8-10}\cmidrule(lr){11-12}
    &Network sparsity&mAP&AP50&Network sparsity&mAP&AP50&Network sparsity&mAP&AP50&Network sparsity&mAP\\
    \cmidrule(lr){1-1}\cmidrule(lr){2-4}\cmidrule(lr){5-7}\cmidrule(lr){8-10}\cmidrule(lr){11-12}
    $90\%$& $41.99\%$& $35.46$& $56.51$& $41.99\%$& $32.40$& $52.89$& $31.49\%$& $62.27$& $84.93$&$65.37\%$&$61.75(\pm0.22)$ \\
    $80\%$& $37.33\%$& $36.52$& $57.28$& $37.33\%$& $33.53$& $54.15$& $28\%$& $63.48$& $85.72$&$58.11\%$& $67.30(\pm0.39)$ \\
    $50\%$& 24.55\%& 37.99& 58.83& 24.55\%& 34.76& 55.91& 19.71\%& 64.21& 86.32&$36.32\%$&$70.32(\pm0.23)$ \\
    $0\%$& $0\%$& $38.5$& $59.29$& $0\%$& $35.13$& $56.39$& $0\%$& $64.59$& $86.48$&0\%&$71.21(\pm0.32)$\\
    \bottomrule
    \end{tabularx}
    
    \vspace{0pt}
    \label{tab:in_res50_coco_mask}
\end{table*}

\section{Error analysis on downstream tasks}
\label{sec:errors}
The mAP score provides us a good way to summarize the performance of an object recognition model with a single number. But it hides a lot of information regarding what kind of mistakes the model is making. Do the sparse subnetworks obtained by LTH make same mistakes as the dense models? In order to answer this question, we consider a dense Mask R-CNN model with ResNet-50 backbone and a sparse Mask R-CNN model with 20\% of the parameters obtained via LTH. Both the models achieve same performance on downstream tasks as also discussed in Section 4.2 of the paper. 

\subsection{Object Detection and Instance Segmentation}

We resort to a toolbox from ~\cite{tide-eccv2020} to analyze object detection and instance segmentation errors. We consider 5 main sources of errors in object detection. (i) `Cls' refers to an error corresponding to miss-classification of a bounding box by a model, (ii) `Loc' refers to the case when bounding box is classified properly but not localized properly, (iii) `Dupe' corresponds to the errors when model makes multiple predictions at the same location, (iv) `Bkgd' are the cases when background portion of the image (with no objects) are tagged as an object, and finally (v) `Missed' cases when the objects are not detected by the model.

\begin{figure}[t]
\centering
\begin{tabularx}{\linewidth}{C{1.0}C{1.0}}
  \includegraphics[width=0.22\textwidth]{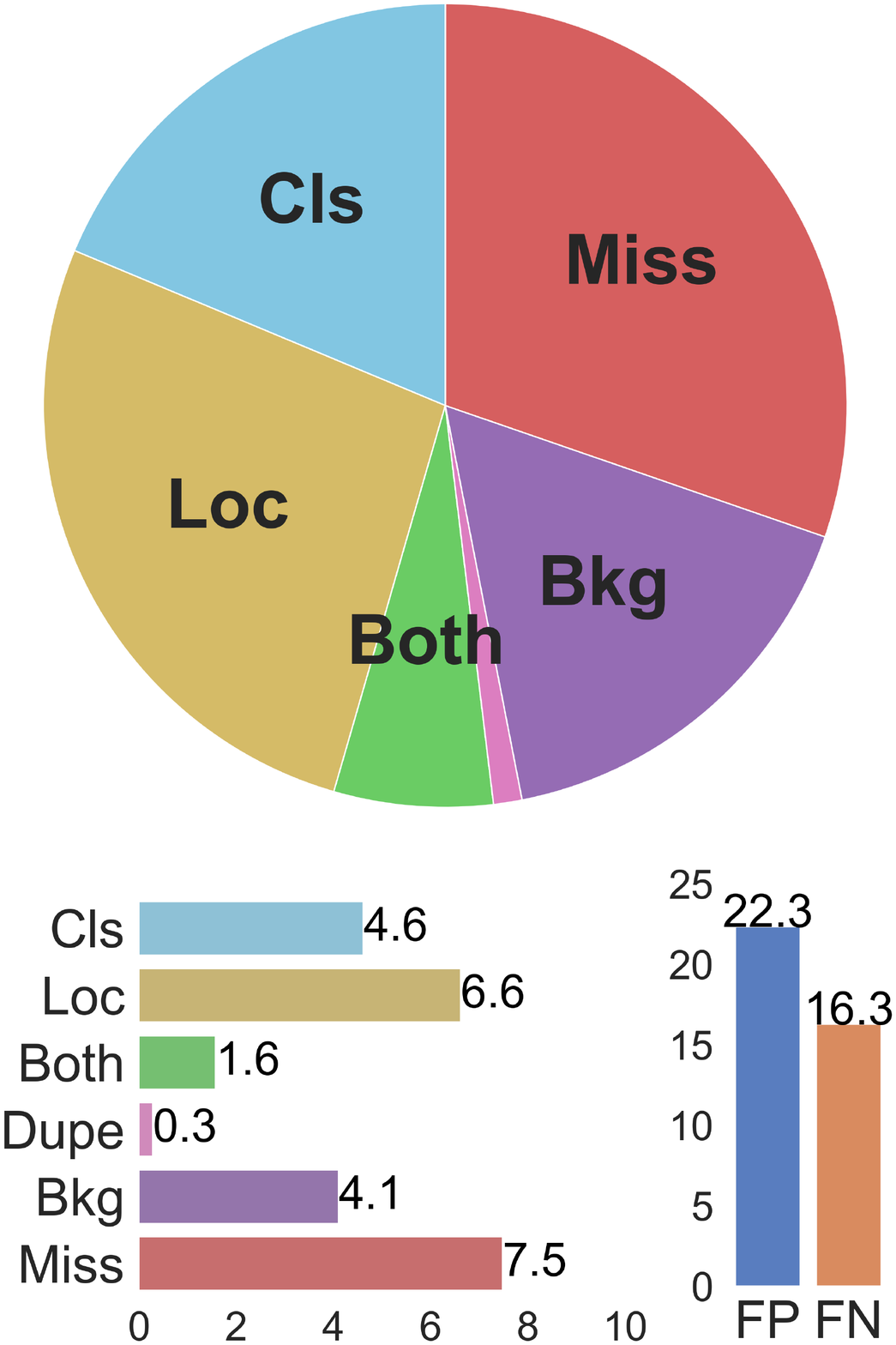} &   \includegraphics[width=0.22\textwidth]{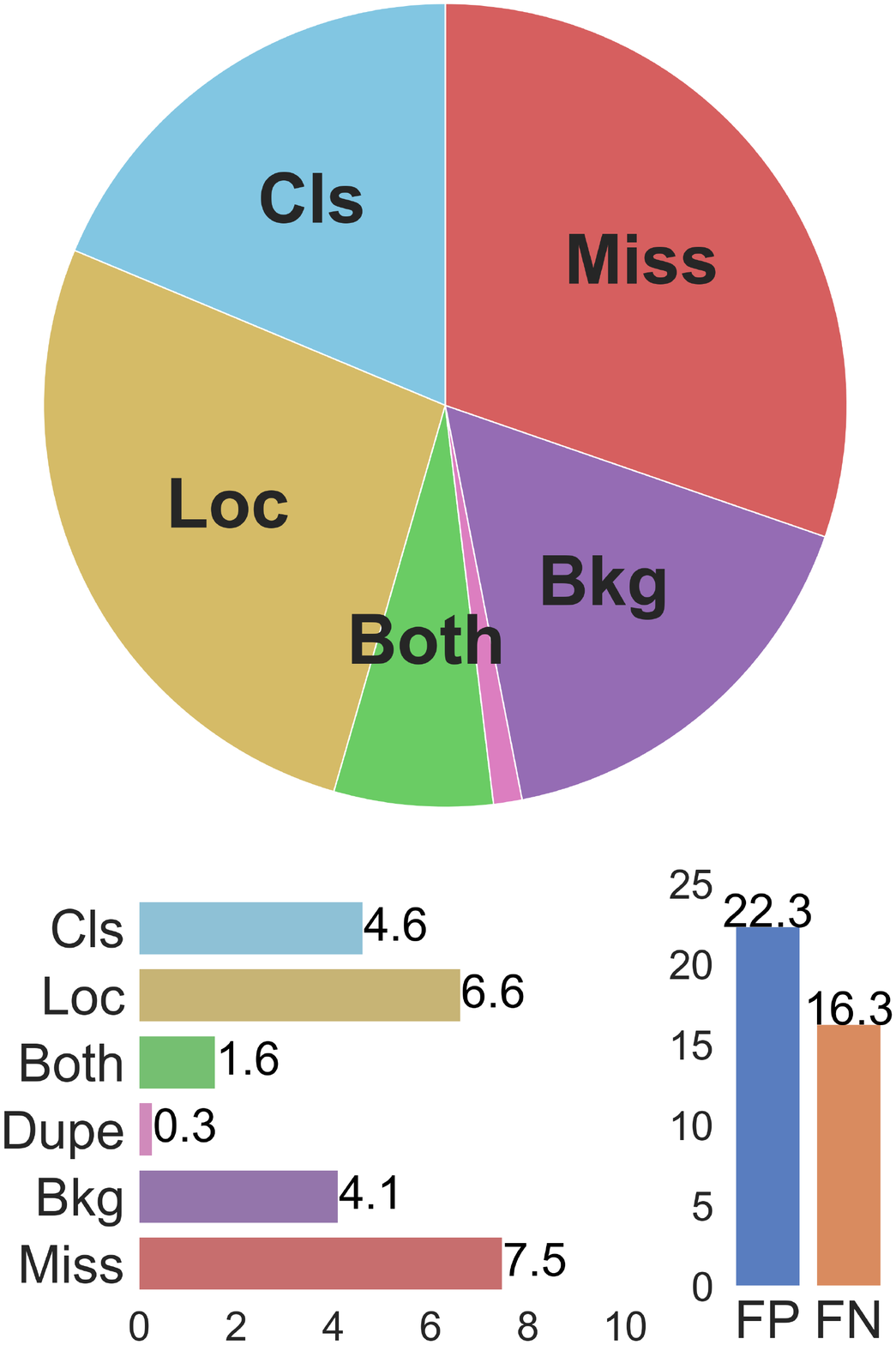} \\
(a) Unpruned model & (b) Pruned model  \\
\end{tabularx}
\caption{Error analysis  of unpruned \vs pruned on object detection. The error types of unpruned and pruned models are nearly the  same.}
\label{fig:box-error}
\end{figure}

\begin{figure}
\centering
\begin{tabularx}{\linewidth}{C{1.0}C{1.0}}
  \includegraphics[width=0.22\textwidth]{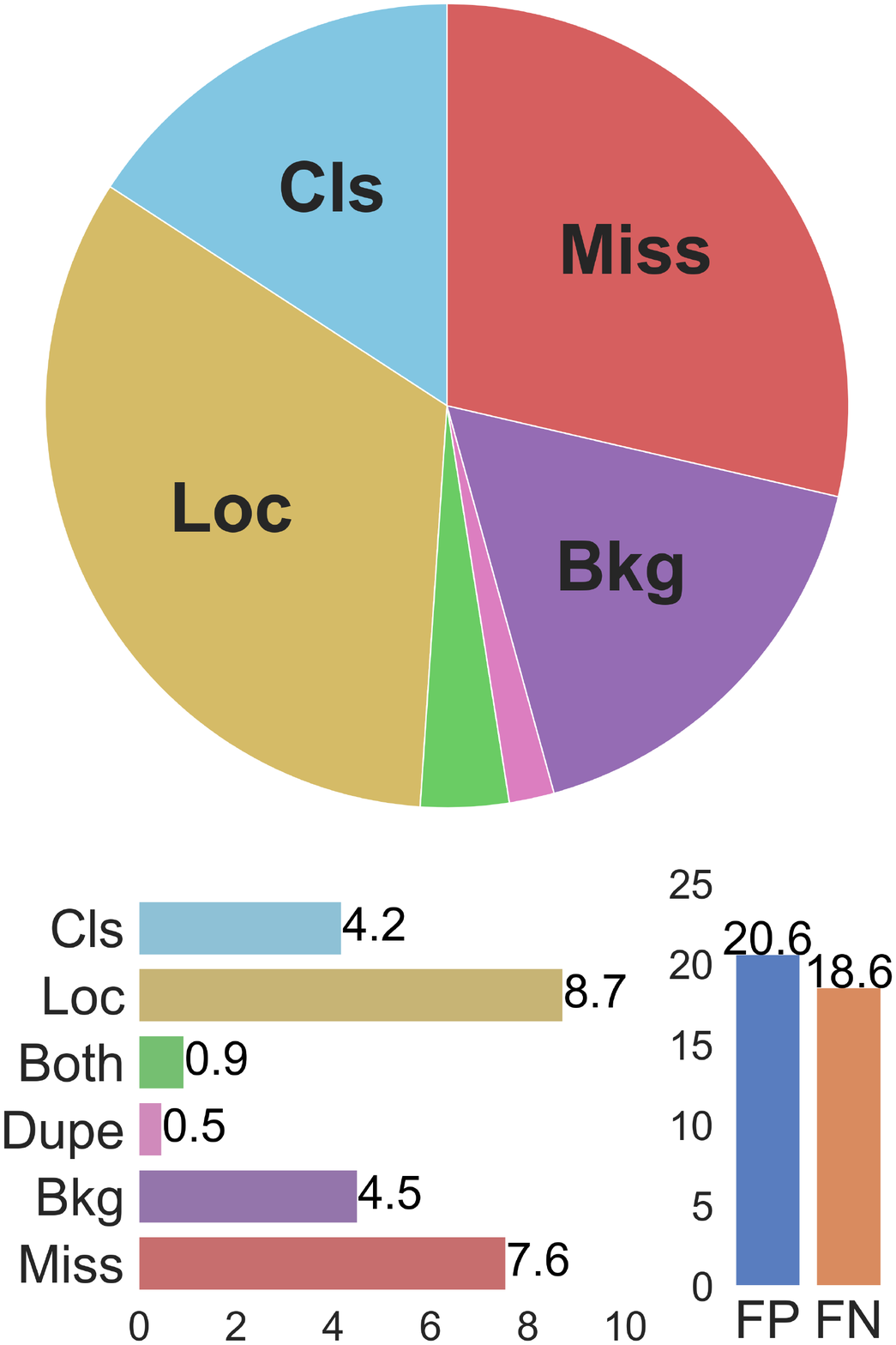} &   \includegraphics[width=0.22\textwidth]{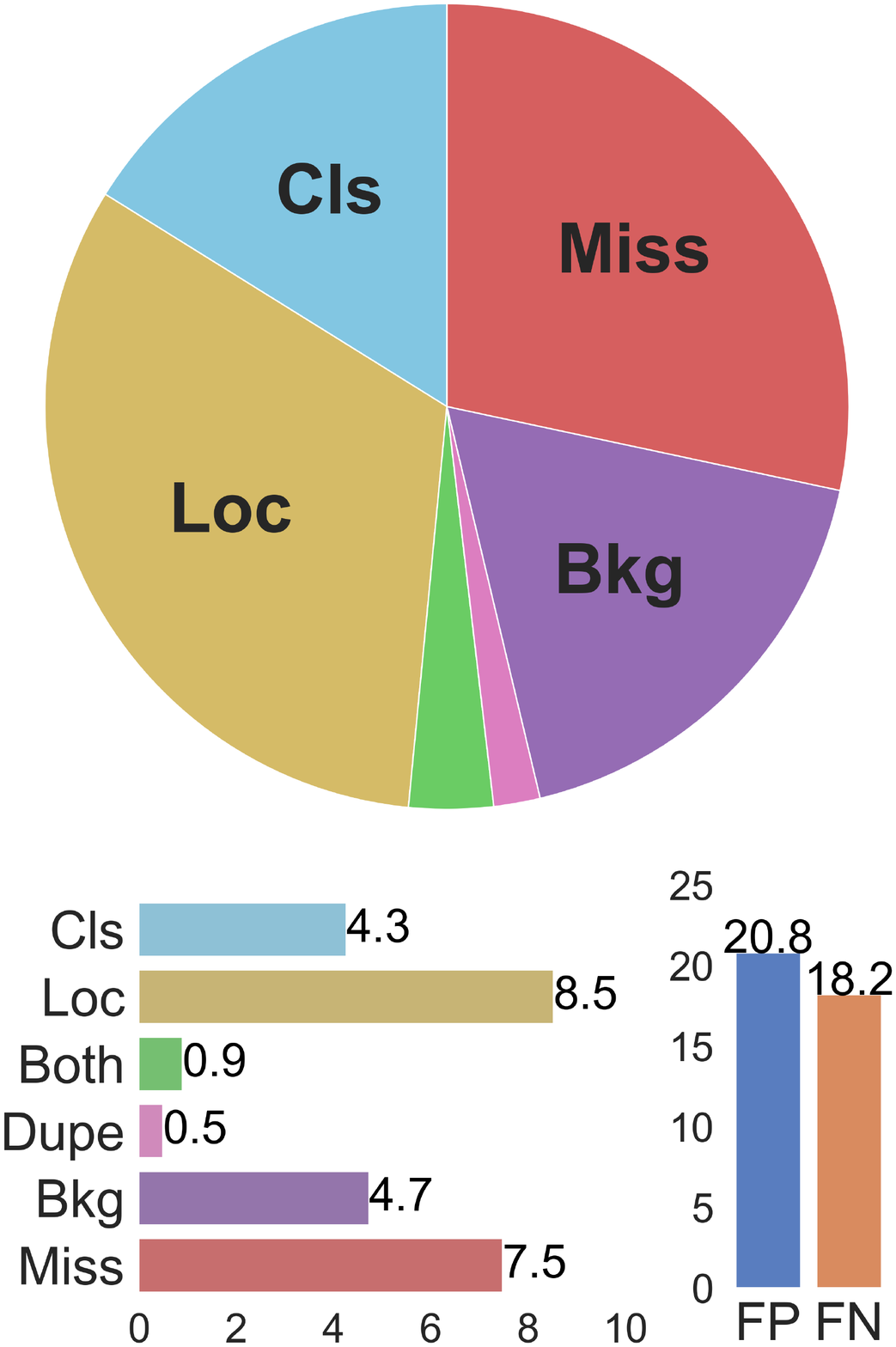} \\
(a) Unpruned model & (b) Pruned model  \\
\end{tabularx}
\caption{Error analysis  of unpruned \vs pruned on instance segmentation.
The error types of unpruned and pruned models are quite similar.}
\label{fig:seg-error}
\end{figure}

Figures~\ref{fig:box-error} and~\ref{fig:seg-error} summarize the analysis of detection and segmentation errors obtained for dense model as compared to a sparse model (with only 20\% of the weights). While in the case of object detection, the performance of both the models is identical, subtle differences emerge in case of segmentation where sparse model makes fewer localization errors but higher background errors.


\begin{figure*}[t]
\centering
\begin{tabularx}{\linewidth}{C{1.0}C{1.0}}
  \includegraphics[width=0.45\textwidth]{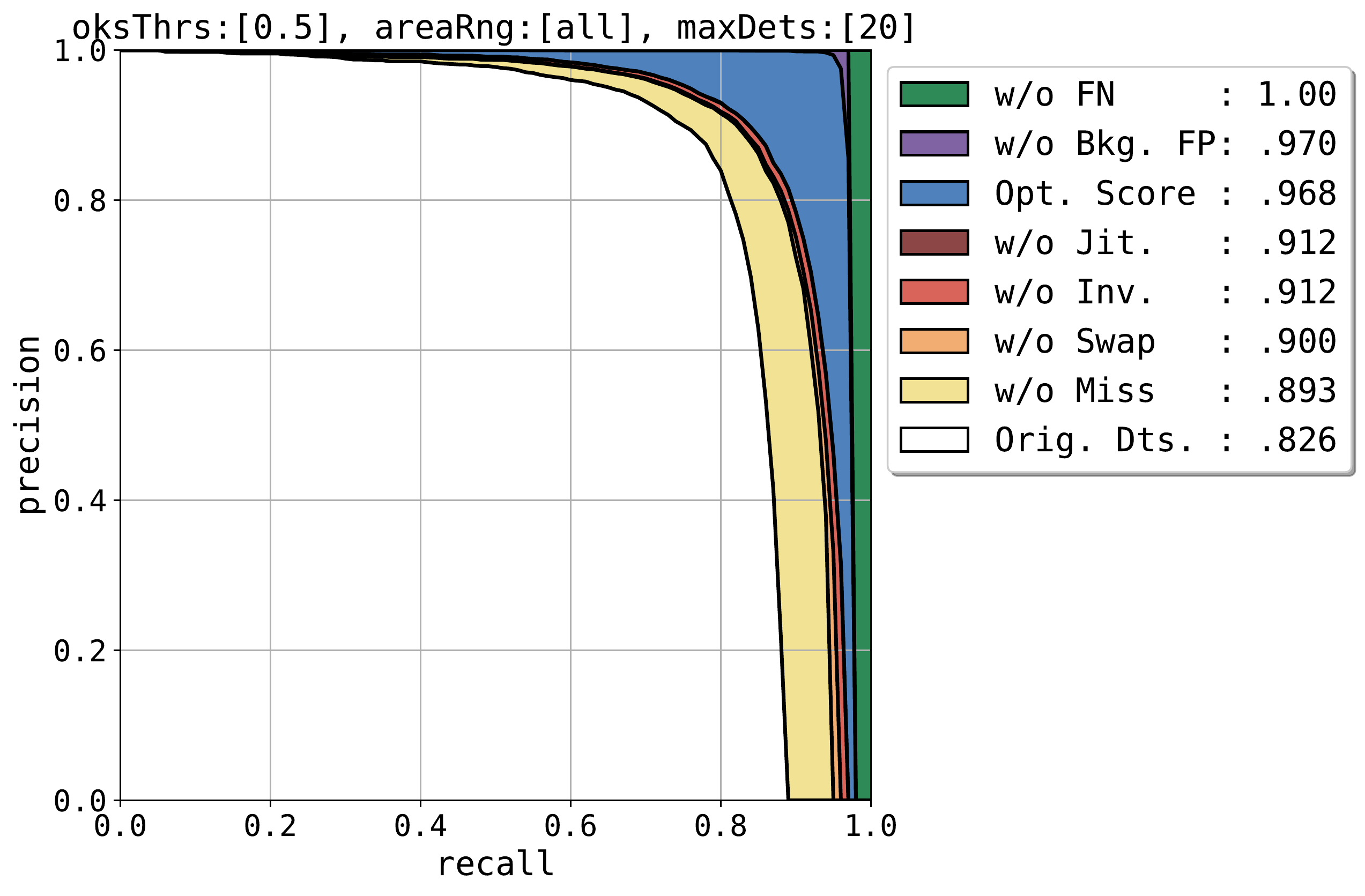} &   \includegraphics[width=0.45\textwidth]{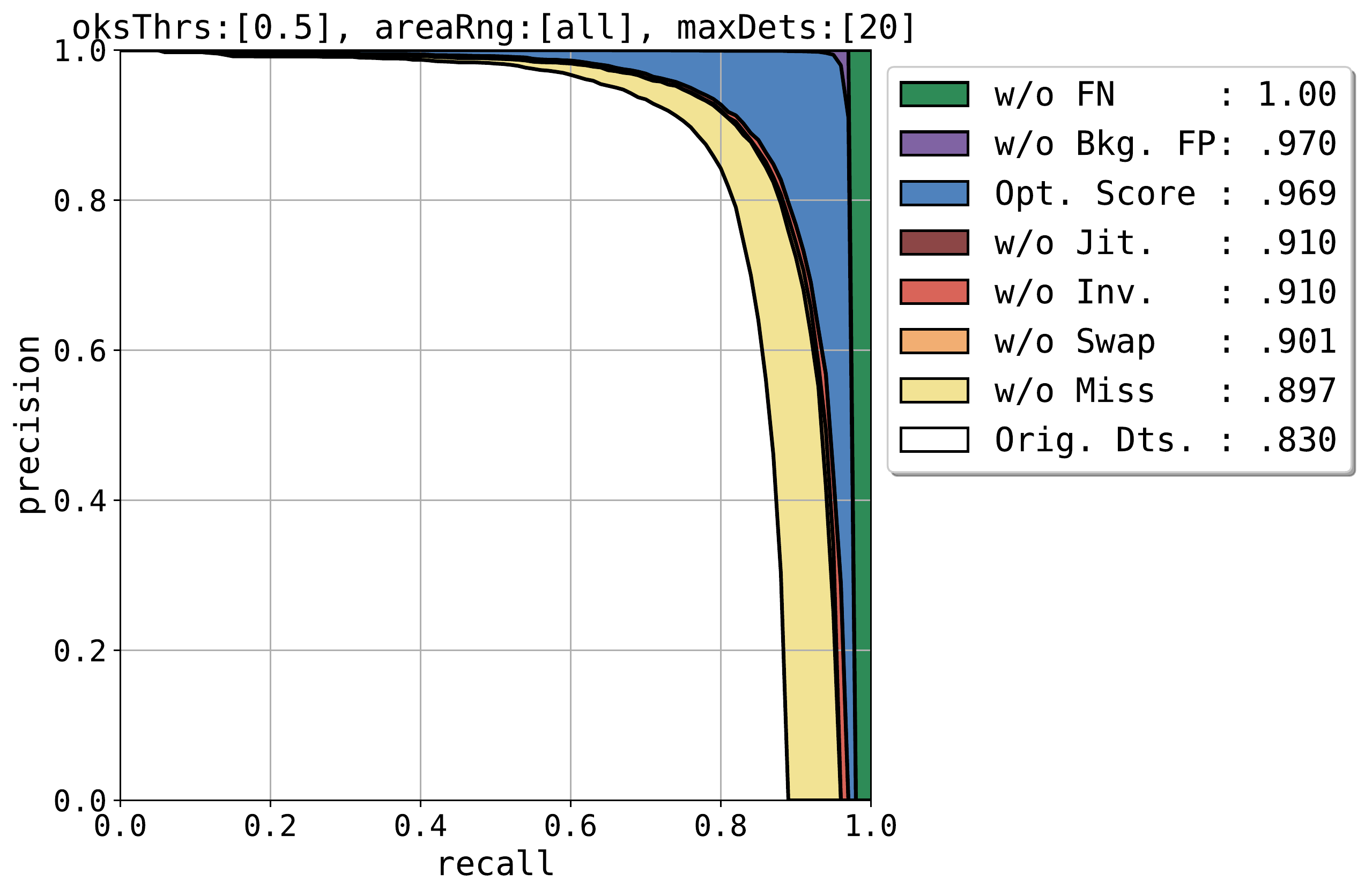} \\
  \includegraphics[width=0.45\textwidth]{figs/props/base_key_50.pdf} &   \includegraphics[width=0.45\textwidth]{figs/props/LTH_key_50.pdf} \\  
(a)Unpruned model & (b) Pruned model  \\

\end{tabularx}
\caption{Error analysis  of unpruned \vs pruned on kyepoint estimation.
The error types of unpruned and pruned models are quite similar while the unpruned one has slightly better performance}
\label{fig:keypoint-error}
\end{figure*}
\begin{figure*}
\centering

\begin{tabularx}{\linewidth}{C{1.0}C{1.0}C{1.0}}
  \includegraphics[width=0.3\textwidth]{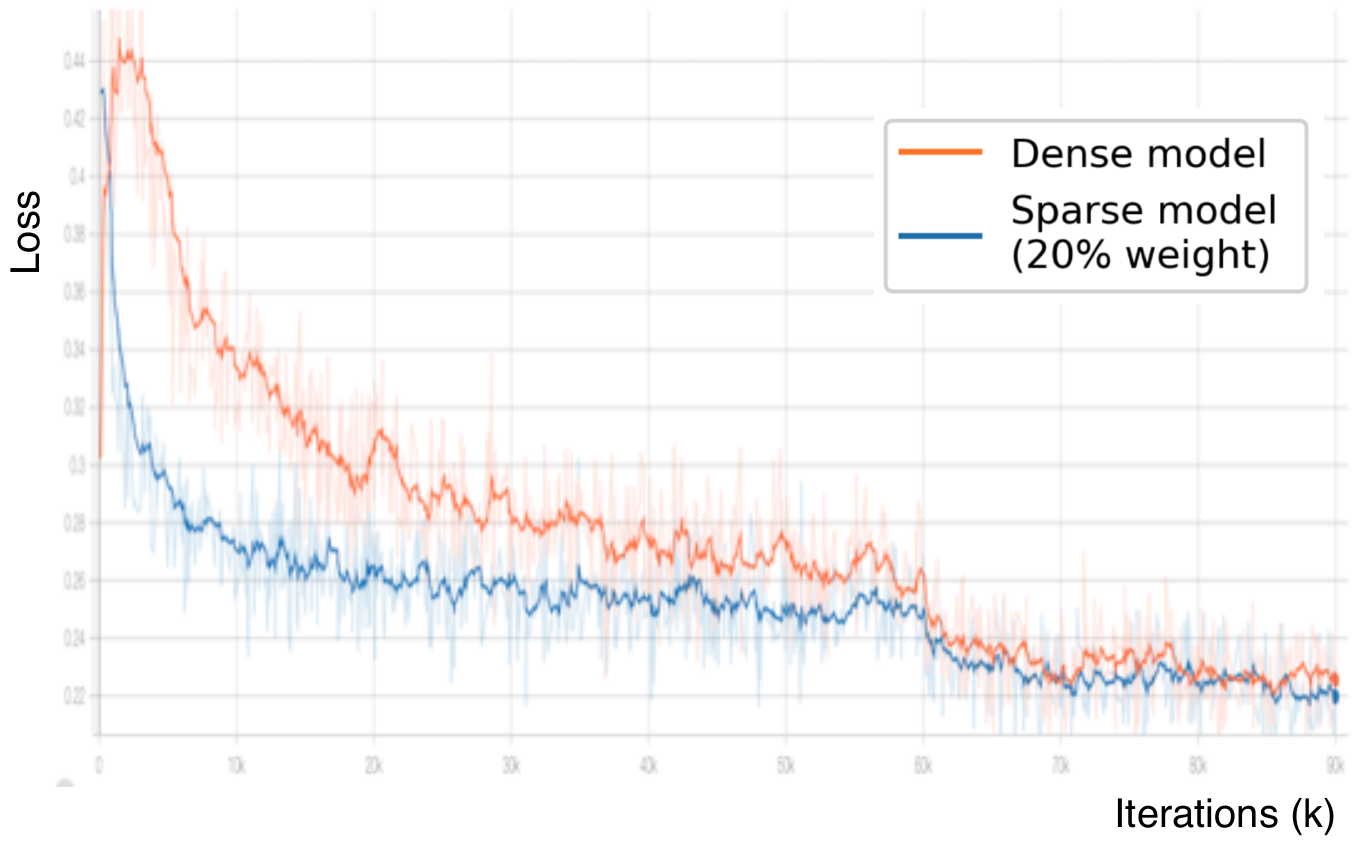} &   \includegraphics[width=0.3\textwidth]{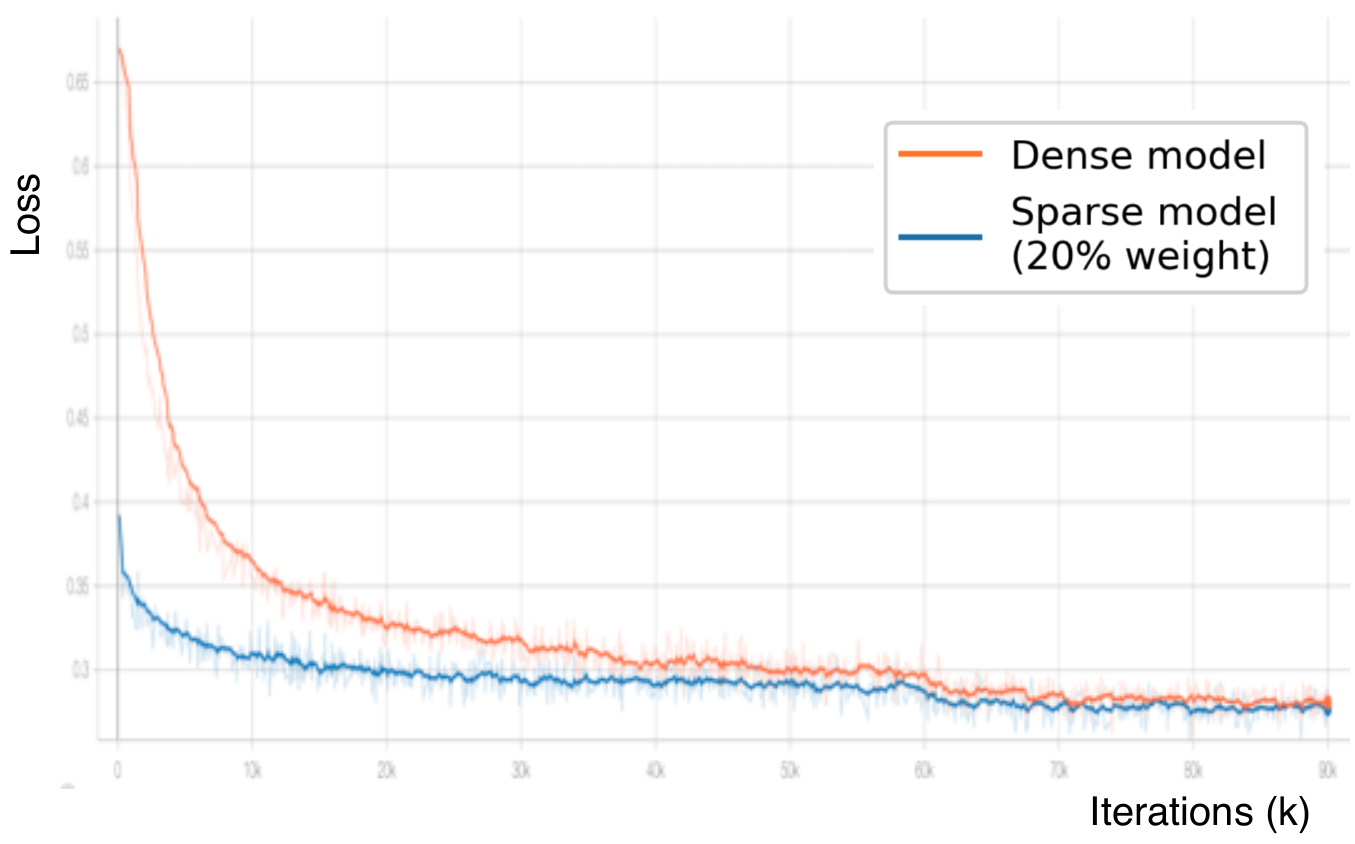} &
 \includegraphics[width=0.3\textwidth]{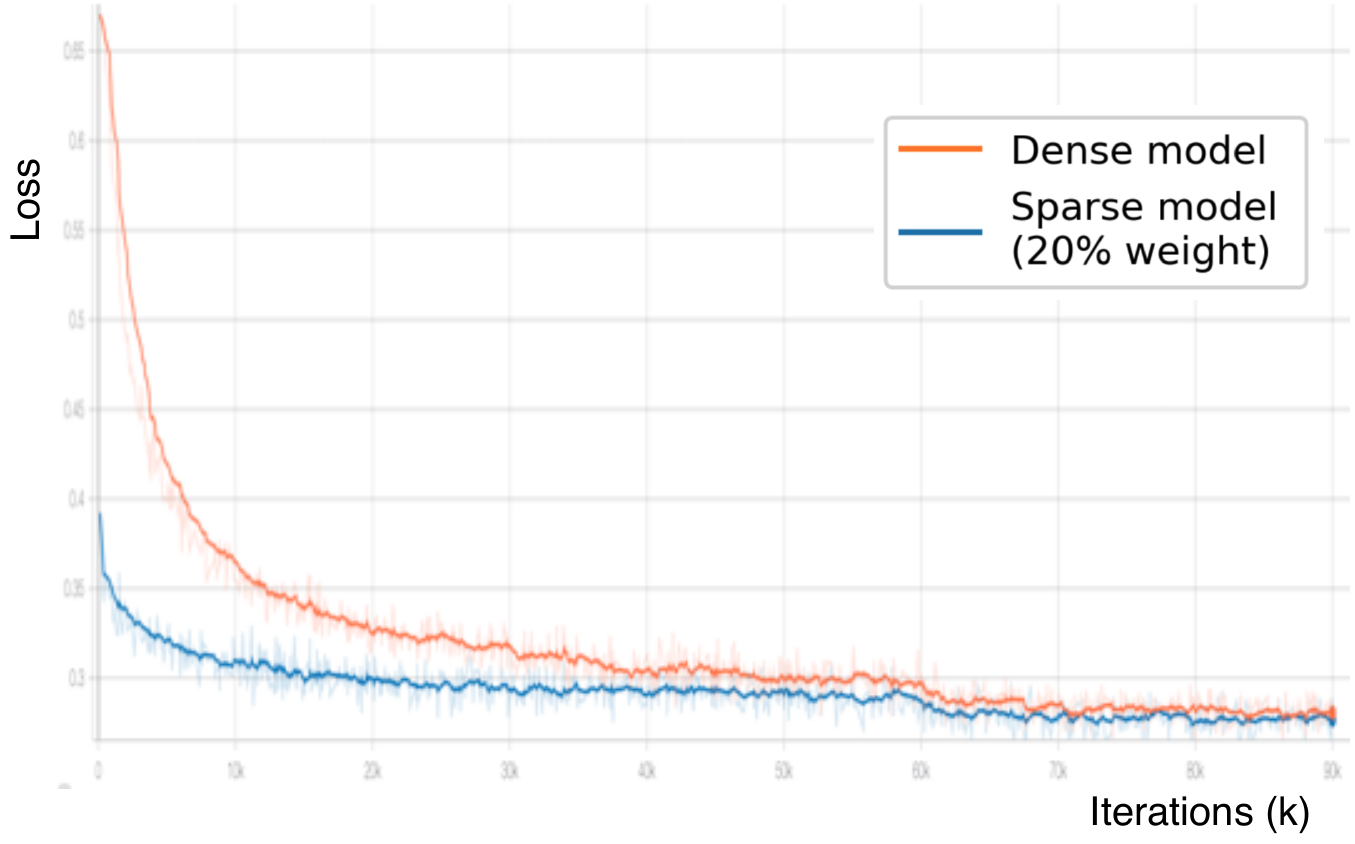} \\
(a) Object detection & (b) Instance segmentation&
(c) Keypoint estimation    \\
\end{tabularx}
\caption{Training curves of dense model \vs sparse model}
\label{fig:train-curve}
\end{figure*}

\subsection{Keypoint Estimation}
We use~\cite{DBLP:conf/iccv/RonchiP17} to perform a similar analyses for sparse and dense models on the task of keypoint estimation. In case of keypoints, we compute the Precision Recall Curve of the model while removing the impact of individual errors of following kinds --- (i) `Miss' - large localization errors, (ii) `Swap' - confusion between same keypoint of two different persons, (iii) `Inversion' - confusion between two different keypoints of the same person, (iv) `Jitter' - small localization error, and (v) `FP' - background false positives. Fig.~\ref{fig:keypoint-error} summarizes the results. As in the previous case, it appears that both the dense and sparse models make similar mistakes.

\section{Sparse subnetworks converge faster}
\label{sec:convergence}
The LTH paper ~\cite{frankle2018the} claimed that sparse subnetworks obtained by pruning, often converge faster than their dense counterparts. In this section we verify the claims of the paper on object recognition tasks and found them to hold true. We plot the validation loss during training for the dense unpruned model, and the sparse subnetwork obtained by keeping only 20\% of the weights of dense model. Both the models achieve a similar mAP after convergence. Fig.~\ref{fig:train-curve} shows the task loss against the number of epochs during training.
The comparisons confirm that the sparse subnetwork initialized from the winning ticket weights converge much faster. This observation is consistent for heterogeneous tasks, \eg, object detection, instance segmentation, and keypoint estimation.

\end{document}